\documentclass{article}

\PassOptionsToPackage{numbers}{natbib}

    \usepackage[preprint]{neurips_2024}

\usepackage[utf8]{inputenc} 
\usepackage[T1]{fontenc}    
\usepackage{hyperref}       
\usepackage{url}            
\usepackage{booktabs}       
\usepackage{amsfonts}       
\usepackage{nicefrac}       
\usepackage{microtype}      
\usepackage{natbib}

\usepackage{graphicx}
\usepackage{amsmath,amssymb,amsfonts}
\usepackage{algorithmic}
\usepackage{textcomp}
\usepackage{subcaption}
\usepackage{tabularx}
\usepackage{color}
\usepackage{xcolor,soul}
\usepackage{algorithm}
\usepackage{multirow}
\usepackage{xspace}
\usepackage{soul}
\usepackage{adjustbox}
\usepackage{enumitem}
\usepackage{svg}  

\usepackage{wrapfig, lipsum, booktabs}
\usepackage{newtxtext} 
\usepackage{pgfplots}
\usetikzlibrary{pgfplots.groupplots}
\pgfplotsset{compat=1.3}
\usepackage{tikz}
\usetikzlibrary{patterns}
\usepackage{pgf-pie}
\usepackage{colortbl}

\newdimen\abovecrulesep
\newdimen\belowcrulesep
\makeatletter
\patchcmd{\@@@cmidrule}{\aboverulesep}{\abovecrulesep}{}{}
\patchcmd{\@xcmidrule}{\belowrulesep}{\belowcrulesep}{}{}

\definecolor{demphcolor}{RGB}{144, 144, 144}
\definecolor{mygray}{gray}{0.4}
\definecolor{lightgray}{rgb}{0.9, 0.9, 0.9}
\newcommand{\demph}[1]{\textcolor{demphcolor}{#1}}

\usepackage{adjustbox}

\definecolor{tabhighlight}{HTML}{e5e5e5}
\definecolor{grey}{RGB}{128,138,135}

\definecolor{oorange}{RGB}{215,122,71}
\definecolor{yyellow}{RGB}{230,185,79}
\definecolor{ppurple}{RGB}{122,30,97}
\definecolor{ggreen}{RGB}{112,173,71}

\definecolor{battleshipgrey}{rgb}{0.3, 0.3, 0.3}
\definecolor{brilliantrose}{rgb}{1.0, 0.33, 0.64}
\definecolor{americanrose}{rgb}{1.0, 0.01, 0.24}
\definecolor{jweigreen}{rgb}{0,0.45,0.24}
\definecolor{bluegray}{rgb}{0.1, 0.1, 0.4}
\definecolor{ao(english)}{rgb}{0.0, 0.5, 0.0}
\definecolor{blanchedalmond}{rgb}{1.0, 0.92, 0.8}
\definecolor{atomictangerine}{rgb}{1.0, 0.6, 0.4}
\definecolor{chocolate(web)}{rgb}{0.82, 0.41, 0.12}
\definecolor{bananayellow}{rgb}{1.0, 0.88, 0.21}
\definecolor{goldenbrown}{rgb}{0.6, 0.4, 0.08}
\definecolor{aliceblue}{rgb}{0.94, 0.97, 1.0}
\definecolor{beige}{rgb}{0.96, 0.96, 0.86}
\definecolor{babyblue}{rgb}{0.54, 0.81, 0.94}
\definecolor{camel}{rgb}{0.76, 0.6, 0.42}
\definecolor{cinnamon}{rgb}{0.82, 0.41, 0.12}

\newcommand{\textvqa}{VQA$^\text{Text}$\xspace}
\newcommand{\vqa}{VQA$^\text{v2}$\xspace}

\newcommand{\mmep}{MME$^{\text{P}}$\xspace}
\newcommand{\seedimg}{SEED$^{\text{I}}$\xspace}

\newcommand{\model}{DeCo\xspace}
\newcommand{\avgpool}{AdaptiveAvgPool\xspace}

\usepackage{ifthen}
\newboolean{showcomments}
\setboolean{showcomments}{true} 
\newcommand{\lean}[1]{\ifthenelse{\boolean{showcomments}}{\textcolor{brown}{\bf \small[#1 - wang]}}{}}

\title{\model: Decoupling Token Compression from Semantic Abstraction in Multimodal Large Language Models}

\author{Linli Yao$^{\dagger}$, Lei Li$^{\ddagger}$, Shuhuai Ren$^{\dagger}$, Lean Wang$^{\dagger}$, Yuanxin Liu$^{\dagger}$, Xu Sun$^{\dagger}$, Lu Hou$^{\S}$\\
$^{\dagger}$National Key Laboratory for Multimedia Information Processing, 
\\School of Computer Science, Peking University\\
$^{\ddagger}$The University of Hong Kong\\
$^{\S}$Huawei Noah's Ark Lab\\
\texttt{\small\{linliyao, shuhuai\_ren, liuyuanxin\}@stu.pku.edu.cn} \quad \texttt{\small{nlp.lilei@gmail.com}}, \\
\texttt{\small{\{lean, xusun\}@pku.edu.cn}} \quad \texttt{\small{houlu3@huawei.com}} \\
}

\begin{document}

\maketitle

\begin{abstract}




The visual projector, which bridges the vision and language modalities and facilitates cross-modal alignment, serves as a crucial component in Multimodal Large Language Models (MLLMs).
However, measuring the effectiveness of projectors in vision-language alignment remains under-explored, which currently can only be inferred from the performance of MLLMs on downstream tasks.
Motivated by the problem, this study examines the projector module by interpreting the vision-language semantic flow within MLLMs. 
Specifically, we trace back the semantic relevance flow from generated language tokens to raw visual encoder patches and the intermediate outputs produced by projectors. 
Our findings reveal that compressive projectors (e.g., QFormer), abstract visual patches into a limited set of semantic concepts, such as objects or attributes, resulting in a ``double abstraction'' phenomenon. Therefore, in MLLMs, this involves a first visual semantic abstraction by the projector referring to pre-defined query tokens (in the vision modality), and a second extraction by the LLM based on text instructions (in the language modality). 
The double abstraction is inefficient in training and will result in cumulative vision semantics deficiency.
To mitigate this issue, we propose the key insight of  ``\textbf{De}couple \textbf{Co}mpression from Abstraction \textbf{(\model)}'', that is compressing the visual token number at the patch level by projectors and allowing the LLM to handle visual semantic abstraction entirely. 
Consequently, we adopt a simple compressor, i.e., 2D Adaptive Pooling, to downsample visual patches in a parameter-free manner. 
Empirical evaluation demonstrates that \model surpasses traditional compressive projectors regarding both performance and efficiency. It achieves performance gains of 0.9\%, 7.1\%, and 2.9\% across the MLLM Benchmarks, Visual Localization, and Open-ended VQA tasks with fewer trainable parameters and faster convergence speed. 
Furthermore, \model preserves vision spatial locality and exhibits robustness across various MLLM configurations, including different vision backbones, image resolutions, and LLMs. 
Our code will be available at \small\url{https://github.com/yaolinli/DeCo}.

\end{abstract}

\begin{figure}[htbp]
    \centering
    \begin{minipage}[c]{0.2\textwidth}
        \centering
        \includegraphics[width=\linewidth]{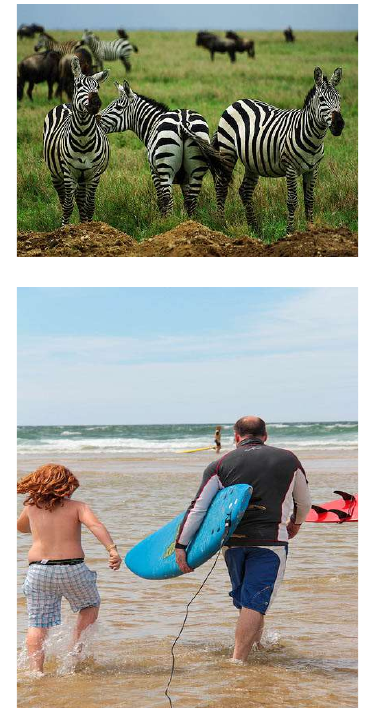} 
        \subcaption{Original Images}
    \end{minipage}%
    \hfill
    \begin{minipage}[c]{0.385\textwidth}
        \centering
        \includegraphics[width=\linewidth]{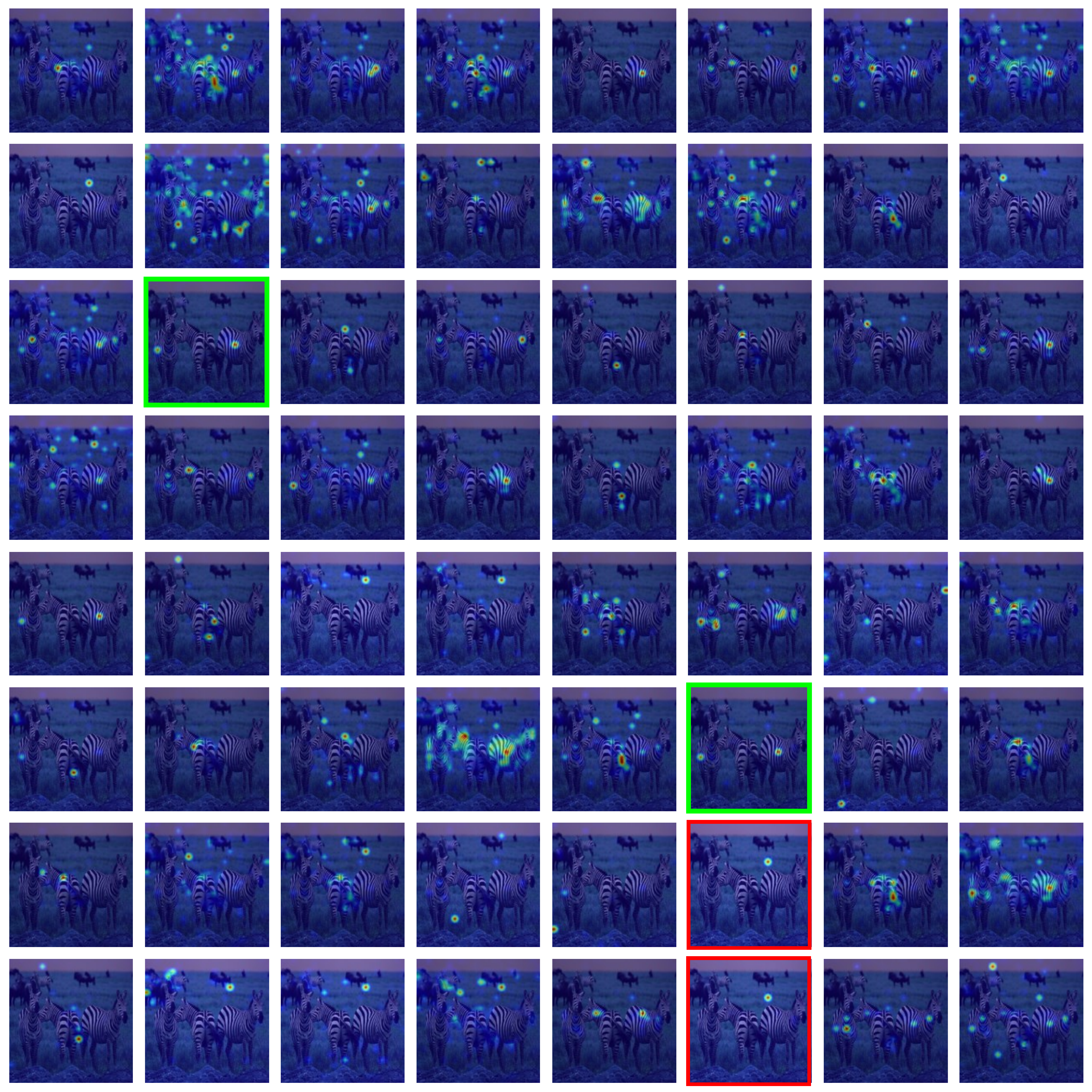} 
        \subcaption{Query-to-Patch Relevance (top img.)}
    \end{minipage}%
    \hfill
    \begin{minipage}[c]{0.385\textwidth}
        \centering
        \includegraphics[width=\linewidth]{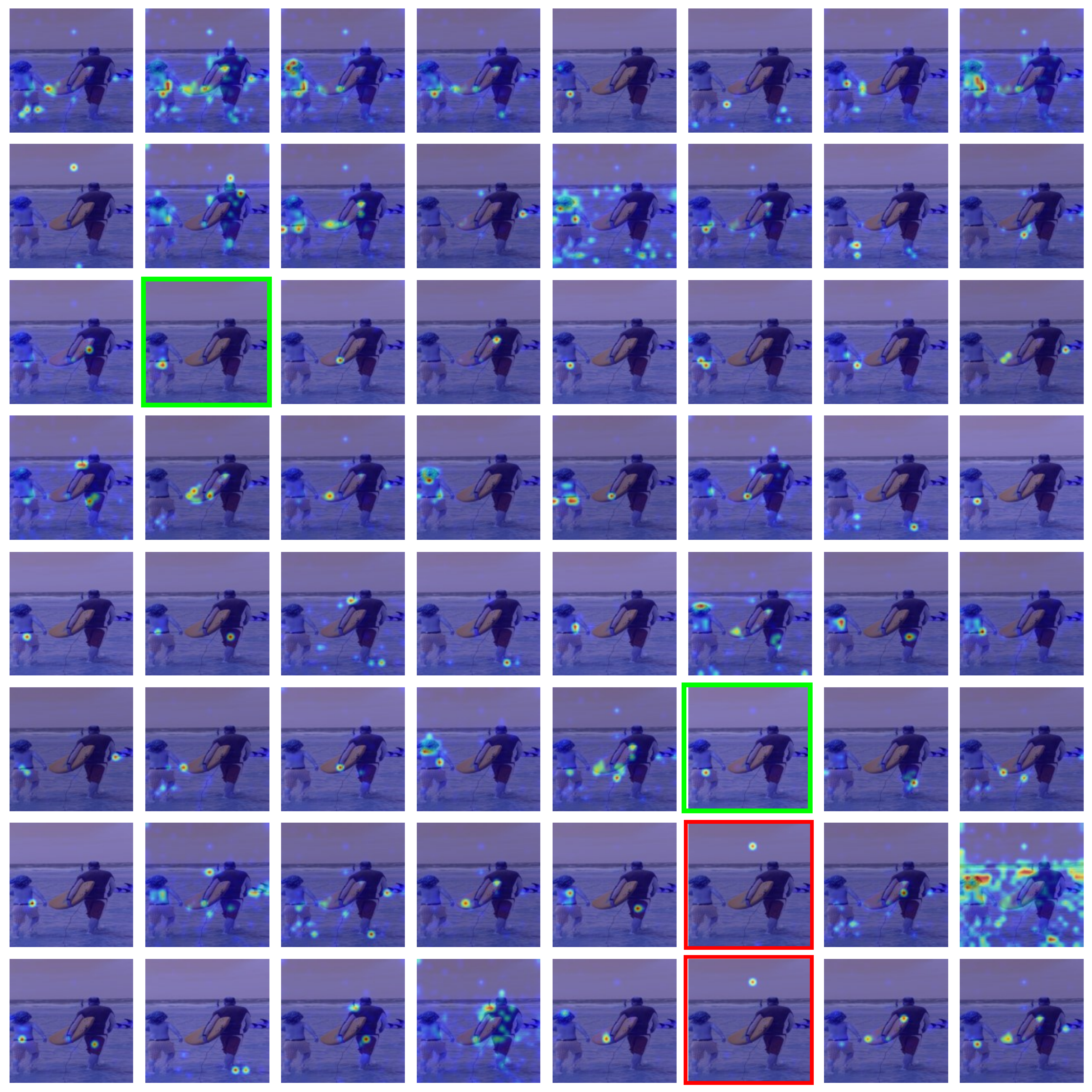} 
        \subcaption{Query-to-Patch Relevance (down img.)}
    \end{minipage}
     \caption{\small 
     Visualization of the R-GAE relevance map from compressed visual tokens (Query) to original image patches (Patch) of the QFormer~\cite{li2023blip2} projector. The QFormer reduces the original $576$ visual (patch) tokens to 64 (equal to $8\times8$) learned query tokens. The relevance maps are obtained from the image-to-text generation process of the MLLM. From the Query-to-Patch map (zoomed in), each query token is activated with diverse visual concepts at the semantic level, such as objects (zebras, grassland, the skateboard), attributes (black and white texture of zebras), and backgrounds (the sea level). However, different query tokens from the same image are visually sparse and showcase repetitive patterns (highlighted in the same color frame), limiting their capacity for visual semantic expression.
     }
     \vspace{-10pt}
    \label{fig:intro_q2i}
\end{figure}

\section{Introduction}
Multimodal Large Language Models (MLLMs)~\citep{gpt4v,gemini,reka2024core} endow Large Language Models (LLMs) with vision perception capability, which have shown their versatility and expertise in diverse vision-language tasks~\cite{kafle2018dvqa, yu2016modeling_refcoco, singh2019text_vqa, vizwiz,li2024multimodal, yao2023edit, yao2022image,chen2023rethinking}. 
For MLLMs, learning good vision-language alignment is at the core of their intelligence~\citep{li2023blip2, zhu2023minigpt4, clip-openness, ren2024prompt}. 
To achieve cross-modal alignment, recent studies utilize an intermediate module, i.e., the projector~\cite{liu2023llava,zhu2023minigpt4,madureira-2021-flamingos,dai2023instructblip}, to map representations of image patches~\citep{dosovitskiy2020image} into the LLM embedding space as visual tokens. 
Widely used projectors can be roughly summarized into two branches: non-compressive and compressive. The non-compressed projector~\cite{liu2023llava} directly uses linear layers that translate the visual token dimension to the LLM's while keeping the visual token number unchanged. 
Despite its simplicity and effectiveness,
the linear projector struggles with high training resources and costs due to the length of the visual token sequence.
The sequence would be long in two common scenarios: (i) the length increases quadratically with the input resolution~\cite{li2023otterhd,internVL}; (ii) the length increases linearly with the image number for handling video frames~\cite{timechat,song2023moviechat,testa},
potentially resulting in sequences up to a million tokens long~\cite{liu2024lwm}.

On the other branch, prevalent compressive projectors, e.g., QFormer~\cite{li2023blip2,dai2023instructblip}, Resampler~\cite{Alayrac2022FlamingoAV}, and D-Abstractor~\cite{cha2023honeybee}, condense the original visual tokens into fewer query tokens to reduce visual redundancy, which have a better balance between performance and efficiency.

However, how existing projectors affect the vision-to-language semantic alignment in an explainable perspective is still under-explored. Understanding this question is crucial for facilitating better architectural improvement and providing broader practicability in demanding scenarios such as high image resolutions and video applications.
In this study, we investigate this problem by analyzing the relevance between generated textual tokens, raw visual patches and intermediate projector outputs. We start by tracing the language-to-vision semantic flow using a novel R-GAE explainability tool. Specifically, we decouple the overall Text-to-Patch semantic relevance to Text-to-Query and Query-to-Patch sub-flows during the image-to-text generation. Among the sub-flows, the Text-to-Patch relevance reveals the effective visual context from ViT~\cite{radford2021clip} (i.e., Patch) leveraged by the LLM (i.e., Text). Meanwhile, the Query-to-Patch relevance interprets the visual patterns learned from original image patches~(i.e., Patch) by query tokens (i.e., Query). 

Based on the R-GAE analysis, we derive two important findings: \textbf{Firstly}, the query tokens \textit{compress} the number of visual tokens by \textit{abstracting} semantic-level visual concepts, leading to visual semantics deficiency such as loss of fine-grained attributes and spatial locality. As Figure~\ref{fig:intro_q2i} illustrates, different query tokens are activated with varied visual concepts such as objects, attributes or backgrounds from the original images. For the top image with zebras in the grassland, query tokens attend to visual patterns such as three zebras, their body parts, surface textures, and distant backgrounds respectively. 
However, the fixed number of queries can only express limited visual semantics. Specifically, different query tokens show repetitive patterns across images (highlighted by color frames in Figure~\ref{fig:intro_q2i}). Moreover, they tend to lose fine-grained visual attributes (e.g., ``purple and red'' in Figure~\ref{fig:model_r-gae}). Furthermore, the vanilla QFormer has been demonstrated to lose visual spatial locality~\cite{cha2023honeybee} during semantic abstraction. 

\textbf{Secondly}, the LLM acts as an excellent visual-semantic abstractor directly from patch features. As Figure~\ref{fig:model_r-gae} first row shows, utilizing a non-compressive linear projector allows the LLM to perceive patch-level visual representations and attend to accurate vision regions without prior semantic deficiency.
Consequently, the QFormer-based MLLM system redundantly extracts the visual semantics twice using QFormer and LLM, which we refer to as the \emph{Redundant Double Abstraction} phenomenon. This double abstraction introduces two major drawbacks: (i) an accumulation of vision deficiencies at the semantic level, meaning that the loss of fine-grained semantics and spatial locality in the QFormer abstraction will propagate to the LLM and (ii) increased training complexity required to optimize an efficient visual semantic abstractor. Therefore, a more efficient compressive projector is needed to simplify the training complexity and preserve more visual context.

In this study, we propose to \textit{\textbf{De}couple token number \textbf{Co}mpression (\textbf{DeCo}) from vision semantic abstraction}. The core of \model is using \textit{a simpler projector, which operates and outcomes visual tokens directly at the patch level} to reduce the visual token number. Subsequently, the LLM independently abstracts vision semantic concepts from the reduced tokens.  
In the DeCo framework, we adopt a simple Adaptive Average Pooling as a natural down-sampler at the patch level and then use the linear layers to map the vision dimension. This projector has three-fold advantages. Firstly,  it can flexibly compress the visual tokens into an arbitrary indicated number by automatically calculating the pooling kernel size and stride. It is also parameter-free and thus converges faster. Besides, it preserves the vision spatial locality via the kernel-based operation and neighbor patch integration. 
Experiments comparing prevailing compressed projectors 
under the same settings verify the effectiveness and efficiency of our DeCo framework. 
Meanwhile, DeCo shows stronger spatial understanding ability, as well as demonstrates robustness across various MLLM configurations, including different vision backbones, image resolutions, and LLMs.

In short, we make three key contributions in this work: (i) We design a novel analysis tool R-GAE to dissect the learned visual semantics in projectors of MLLMs. 
(ii) Using this tool, we reveal a double abstraction phenomenon at semantic level, which leads to deficiencies in MLLM performance. To address this, we propose DeCo, a simple architecture with an adaptive pooling mechanism to decouple token compression and visual semantic abstraction. (iii) Experimental results demonstrate that our DeCo framework is simple yet effective, significantly boosting the spatial understanding capabilities of MLLMs across various benchmarks.

\vspace{-5pt}
\section{Related Work}
\vspace{-5pt}
\paragraph{Multimodal Large Language Models.}
The development of large vision-language models has accelerated recently~\citep{gpt4v,reka2024core,gemini,2023vlfeedback}. Flamingo~\citep{Alayrac2022FlamingoAV,awadalla2023openflamingo} and IDEFICS~\citep{laurencon2023obelics} have showcased the effectiveness of consolidating LLMs with vision encoders. The Q-Former from BLIP-2~\citep{li2023blip2} has helped bridge the gap between the visual and text modalities. InstructBLIP~\citep{dai2023instructblip}, Ying-VLM~\citep{li2023m3it} and MM-ICL~\citep{zhao2023mmicl} further integrate instructions into the vision-text alignment process for improved in-context learning ability~\citep{icl_survey}.
Various approaches have been proposed to align visual encoders and LLMs effectively. MiniGPT-4~\citep{zhu2023minigpt4} and LLaVA~\citep{liu2023llava,liu2023llava15} use a single projection layer, while mPLUG-Owl~\citep{ye2023mplugowl} adopts LoRA tuning~\citep{hu2021lora,ma2024paramtuning}, showing promising results. Qwen-VL-Chat~\citep{Qwen-VL} has scaled up multi-modal pre-training with more datasets. Fuyu-8~\citep{fuyu-8b} proposes a new architecture by segmenting images into pixel patches, treating them as visual tokens to train a conditional multi-modal language model directly. 
However, these works employ projector modules empirically or simply refer to the final performance of the MLLMs on downstream tasks without conducting an in-depth analysis of the projectors' effectiveness.
In this paper, we examine this significant component by tracking the vision-and-language semantic flow within MLLMs. We visualize the internal patterns learned by projectors and highlight their drawbacks, offering valuable insights for future development.
\vspace{-5pt}

\paragraph{Transformer Explainability.}
Explainability tools have been widely explored for Transformers to better visualize their inner decision-making processes. Raw attention maps in Transformers usually provide interpretations for a single layer. Abnar et al.~\cite{abnar2020quantifying} combine the attention scores across multiple layers and propose the rollout method. Chefer et al.~\cite{chefer2021transexplain} introduce the relevance map through information propagation from all layers and components in Transformers. LRP~\cite{voita2019lrp} captures the relative importance between different attention heads using gradients.  Casual Interpretation~\cite{rohekar2024causal} can identify the most important input tokens corresponding to the model output. However, these methods are only applicable to  Transformers with self-attention layers. As an alternative, the GAE~\cite{chefer2021gae} method extends the propagated relevance maps to bi-modal scenarios with cross-attention layers. Moreover, several studies~\cite{aflalo2022vl, liu2023multimodal, lyu2022dime, ramesh2022investigation, swamy2024multimodn} focus on the multimodal system interpretation. Recently, LVLM-Interpret~\cite{ben2024lvlm-interpret} has developed an interactive application for interpreting  MLLMs. Despite these efforts, in-depth explainability of existing MLLMs is rarely explored. In this study, we propose the R-GAE method derived from GAE for MLLMs to investigate how projector modules affect the vision-and-language semantic alignment of MLLMs.

\section{Visual Projector Analysis}
\label{sec:analysis}

\begin{figure}[t]
    \centering
    \includegraphics[width=\linewidth]{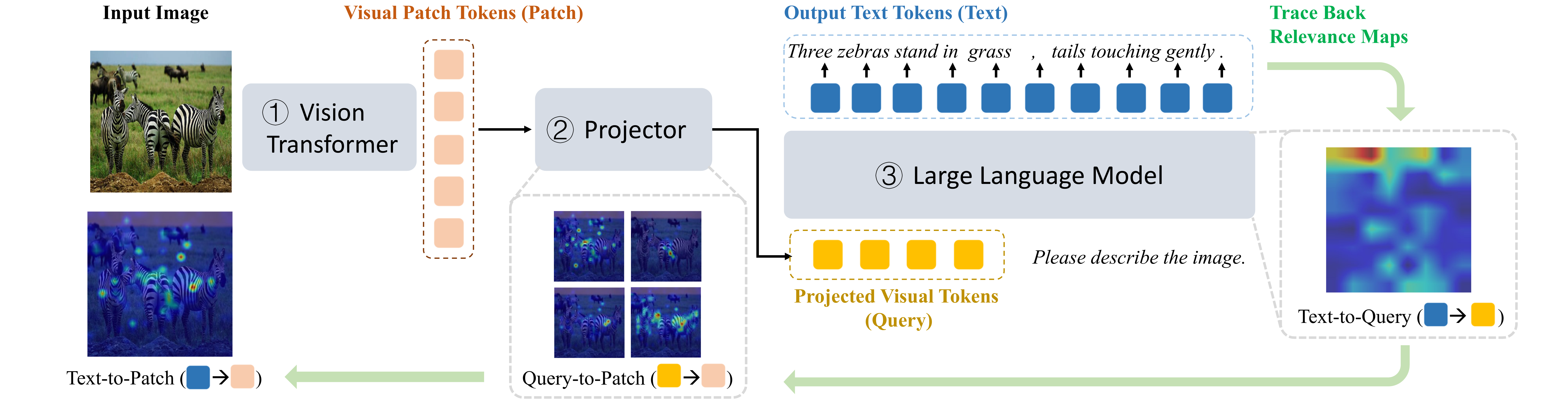}
    \caption{The overall analysis framework of a typical MLLM. During image-to-text generation, we trace back the language-to-vision semantic flow utilizing R-GAE relevance maps. }
    \vspace{-10pt}
    \label{fig:supp_overall}
\end{figure}

\begin{figure}[tbp]
    \centering
    \includegraphics[width=\linewidth]{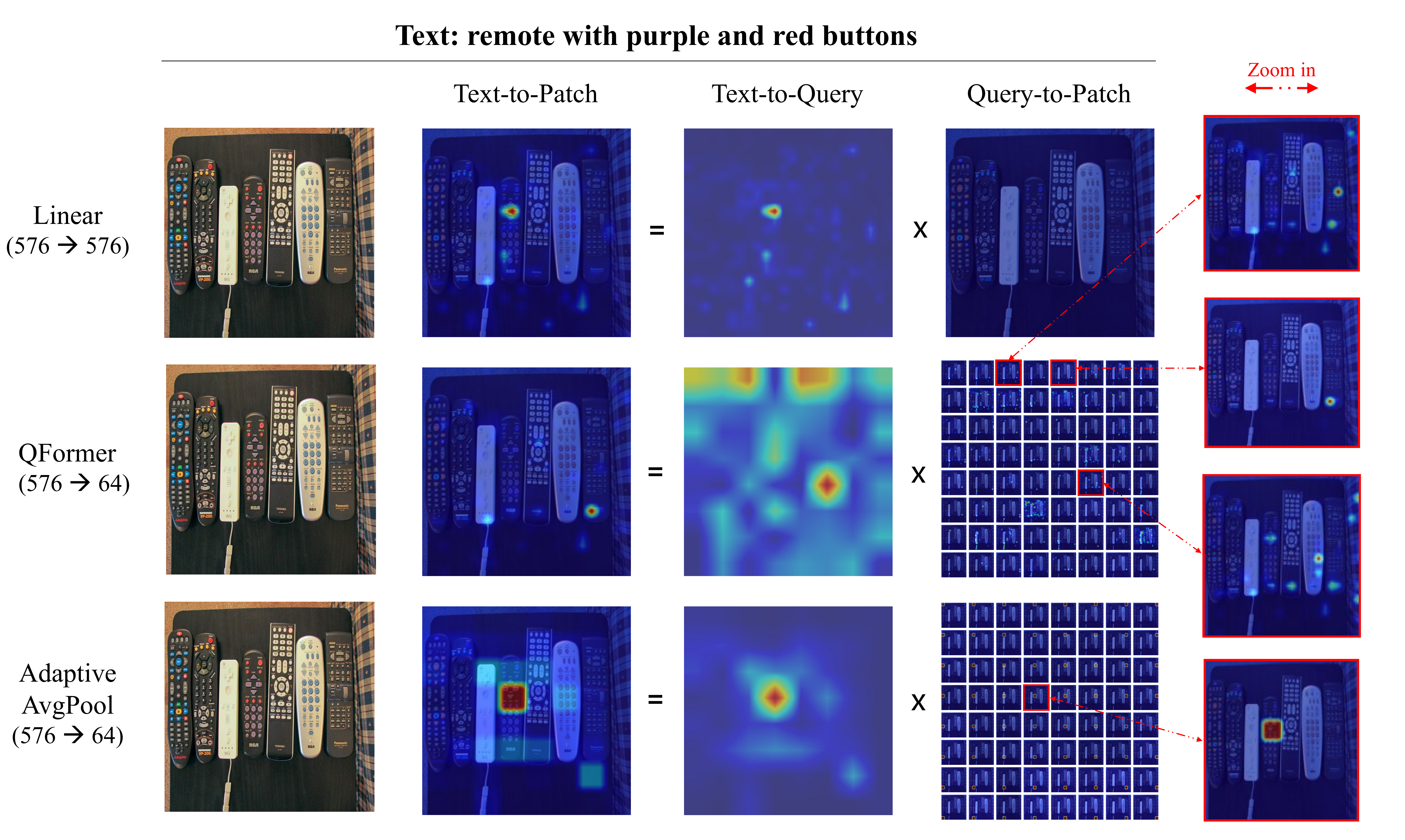}
    \vspace{-10pt}
    \caption{\small Visualization of the R-GAE relevance maps across the same MLLM architecture except for projector modules. The linear projector is non-compressive while the QFormer and Adaptive Average Pooling (ours) compress the original 576 vision tokens to 64 tokens. Text-to-Patch relevance reveals the effective vision semantics aligned with the LLM during image-to-text generation. 
    For QFormer in the second row, its Query-to-Patch map discards the fine-grained visual semantics about ``purple and red''. This semantic deficiency is transmitted to the final Text-to-Patch map and leads to a misalignment of vision patches and textual words.
    }
    \vspace{-5pt}
    \label{fig:model_r-gae}
\end{figure}

In this section, we analyze the impact of projector modules in Multimodal Large Language Models (MLLMs) from a semantic flow perspective using a novel R-GAE explainability tool.
During image-to-text generation, visual context plays an indispensable role in the perception of Large Language Models (LLMs). The related relevance maps between image and text, such as attention maps~\cite{vaswani2017attention}, can serve as an interpretation of the vision-language semantic alignment~\cite{chefer2021transexplain, xu2015show-attend-tell, carion2020detr, ren2021iais}.
As Figure~\ref{fig:supp_overall} shows, given an oracle description in the MLLM architecture, the backtracking relevance map from text words to visual patches (referred to as Text-to-Patch) exhibits the visual semantics aligned with the LLM and further indicates the effective visual context leveraged by the LLM.
To examine the impact of projectors as the intermediate module, we dissect the Text-to-Patch relevance map into Text-to-Query and Query-to-Patch sub-maps, as illustrated in Figure~\ref{fig:model_r-gae}. The Query-to-Patch map can explain the visual patterns learned by the query (or compressed) tokens, while the difference between Text-to-Patch and Text-to-Query, exerted by the projector, reveals its impact on the vision-language semantic alignment.

A typical MLLM architecture comprises a Vision Transformer (ViT) to acquire visual representations $\mathcal{I} \in \mathbb{R}^{N \times d_{I}}$  containing $N$ patches, a projector to transform visual representations into the textual embedding space, and an LLM that handles both vision and instruction tokens to output hidden states $\mathcal{T}  \in \mathbb{R}^{L \times d_{T}}$ and generate responses $Y = \{y_1, y_2, \dots, y_L\}$. We summarize widely adopted projectors into two branches:

\textit{Non-compressive Projectors} maintain the number of patch tokens $N$ and only transform the visual embedding dimension to match the dimension of the LLM, as exemplified by the linear projector~\cite{liu2023llava}. The projected visual tokens can be denoted as $\mathcal{Q}  \in \mathbb{R}^{N \times d_{T}}$. 

\textit{Compressive Projectors} reduce the number of patch tokens  $N$ to a specified lesser number $M$ ($M<N$), conserving training resources. For instance, QFormer~\cite{li2023blip2} learns pre-defined query tokens to compress original visual tokens. These compressed query tokens  $\mathcal{Q}  \in \mathbb{R}^{M \times d_{T}}$ 
 are then fed into the LLM providing vision information.
 
For clear clarification, we distinguish the \textit{compression}  and \textit{abstraction} concepts in this study. The \textit{compression} refers to the reduction of vision token number in particular, whereas \textit{abstraction} denotes the extraction of vision semantic concepts (e.g., objects and attributes, etc.).

\subsection{R-GAE: Relevance Maps in MLLMs Derived from GAE }
\label{subsec:rgae}
We aim to employ the dissected Text-to-Query and Query-to-Patch relevance maps to examine the projector module. A straightforward attempt is utilizing the raw attention maps in MLLM layers as the relevance map~\citep{ren2021iais}. However, the attention map exhibits the interaction between tokens in a single layer~\cite{chefer2021transexplain}. 
Instead, we require a relevance map that traces back inter-token alignment in arbitrary two layers in the MLLM, for instance, the alignment from intermediate-layer query tokens to initial-layer input patch tokens. To achieve this goal, we propose a novel R-GAE relevance map derived from the Generic Attention Explainability (GAE)~\cite{chefer2021gae}. R-GAE extends the GAE method originally designed for classification tasks, to generative MLLMs, and adapts it to the typical MLLM architecture consisting of a ViT, a projector, and an LLM. The R-GAE can acquire relevance maps from any two arbitrary layers within the MLLM through propagation.

We initialize three R-GAE relevance maps including a Text-to-Patch map as $\mathbf{R}_{\mathcal{T}\rightarrow{}\mathcal{I} } $, a Text-to-Query map as $\mathbf{R}_{\mathcal{T}\rightarrow{}\mathcal{Q} } $, and a Query-to-Patch map as $\mathbf{R}_{\mathcal{Q}\rightarrow{}\mathcal{I} } $. Each map is an identity matrix based on the intuition that each input token’s relevance score is equal in the beginning.
Given an image and an instruction (e.g., ``\textit{Please describe the image with a concise sentence}''), an MLLM will generate a textual description $Y = \{y_1, y_2, \dots, y_L\}$ referring to the visual information. During the generation step $t$, we can cache the attention maps across the ViT, the projector and the LLM during a forward pass. Then, specifying a word class $\widehat{y}_t$ as the target prediction,  we can obtain the related gradients through a backward pass. For each layer, a single R-GAE relevance map is obtained by utilizing gradients to average across the attention heads. For step $t$, we can propagate the Text-to-Query map $\mathbf{R}^t_{\mathcal{T}\rightarrow{}\mathcal{Q} } \in \mathbb{R}^{1\times M}$ from the LLM's first layer to its last layer to get the final map. Similarly, the Query-to-Patch map $\mathbf{R}^t_{\mathcal{Q}\rightarrow{}\mathcal{I} } \in \mathbb{R}^{M\times N}$ can be propagated from the first layer to the last layer of the projector. Subsequently, the overall Text-to-Patch relevance map can be obtained by matrix multiplication of Text-to-Query and Query-to-Patch maps:
\begin{align}
    \mathbf{R}^t_{\mathcal{T}\rightarrow{}\mathcal{I} }&=
    \mathbf{R}^t_{\mathcal{T}\rightarrow{}\mathcal{Q} } 
    \times \mathbf{R}^t_{\mathcal{Q}\rightarrow{}\mathcal{I} }
\end{align}

For a complete sentence $Y$, we integrate the R-GAE relevance maps from each time step $t$ by averaging to obtain the overall visual relevance related to a factual sentence. 
We set the ground-truth description from an image-text pair as the target response to perform the backward process. This limits MLLMs with different projectors having the same Oracle Text-to-Patch visualization.
We provide the background of GAE and the specific propagation formula of R-GAE in Appendix A. Moreover, we compare the visualization between R-GAE and original attention maps in Appendix B.

\subsection{A Redundant Double-Abstraction Phenomenon Resulting from  Compressive Projectors }
\label{subsec:projector_analysis}
Based on the R-GAE maps, we analyze the different types of projectors and investigate how they affect the vision-to-language semantic alignment. For fair comparison and analysis, we train MLLMs under the same architecture, except for the projector module, and keep all other variables the same (experimental details are provided in \textsection~\ref{subsec:exp_impl}). We visualize the R-GAE maps of a non-compressive projector (i.e., linear layers) and a compressive projector (i.e., QFormer) in Figure~\ref{fig:model_r-gae} and draw the following findings.

\textit{\textbf{Observation 1.} LLMs are good visual semantic abstractors directly from patch representation.}

The non-compressive projector directly inputs the patch representation to the LLM. As shown in the first row of Figure~\ref{fig:model_r-gae}, given a description containing visual objects (i.e., the remote and buttons) and attributes (i.e., purple and red), the LLM can highlight the most relevant visual regions in a fine-grained manner, as it discriminates the accurate remote with purple and red buttons among other similar remotes. This indicates that the LLM has built a strong alignment between textual and visual semantics based on the patch representation. The recent success of MLLMs~\cite{llavanext, li2023otter, chen2023shikra} with non-compressive projection further demonstrates that the LLM itself is an efficient visual semantic abstractor. For instance, LLaVA-Next~\cite{llavanext}, which employs a simple Multi-layer Perceptron (MLP), achieves state-of-the-art performance across diverse multimodal benchmarks.

\textit{\textbf{Observation 2.} Compressive projectors extract limited visual semantic concepts from patches.}
Compressive projectors like QFormer pre-extract visual semantic concepts from patches and provide reduced visual tokens at the semantic level to the LLM. As the Query-to-Patch map in Figure~\ref{fig:model_r-gae} shows, the compressed 8x8 query tokens are activated with visual semantic patterns such as different remotes, buttons, control panels, and the black background board. However, the fixed number of query tokens can only cover limited visual semantic concepts from the image. Comparing the visual patterns among 64 tokens, we find that they are visually repetitive and semantically sparse. 
For instance, query tokens indexing $(0,1)$ and $(2,0)$ are nearly identical and all attend to the bottom-right panel of the right remote. These sparse query tokens lead to a deficiency in visual semantics, losing the fine-grained attribute of ``purple and red buttons''. Consequently, the LLM suffers from this irreversible visual semantic deficiency when re-extracting visual context in the query semantic space. As the Text-to-Query map shows, the LLM primarily attends to the query tokens indexing $(0,2)$, $(0,4)$, and $(4,5)$ (framed in red), resulting in a misalignment of text words and patches verified in the Text-to-Patch map. More visualization cases are presented in Appendix D.

\textit{\textbf{Insight.}  An inefficient MLLM system due to the double abstraction of visual semantics.} 

Based on these observations, we conclude that existing compressive projectors, which learn a fixed number of query tokens, are inefficient compressors for reducing the number of vision tokens. They result in a ``Double Abstraction'' MLLM system, where visual semantics are first abstracted by projectors and then re-extracted by the LLM. This dual-abstraction procedure has two main shortcomings:
(i) Accumulative visual semantics loss. The projector serves as an intermediate module bridging the ViT and LLM, therefore, the visual semantics lost during the initial abstraction by the projector become a bottleneck for the MLLM system.
(ii) Increased training complexity. Optimizing a projector to be an effective semantic abstractor is essential for alleviating semantic loss; however, this increases the training cost and complexity. For instance, Qwen-VL-7B~\cite{Qwen-VL}, which uses a resampler projector, requires 1.4B pretraining and 50M fine-tuning data across three training stages.

\section{\model: Decoupling Vision Token Compression}
\label{sec:model}

\begin{wrapfigure}{r}{0.5\textwidth}
    \vspace{-15pt}
    \centering
    \includegraphics[width=\linewidth]{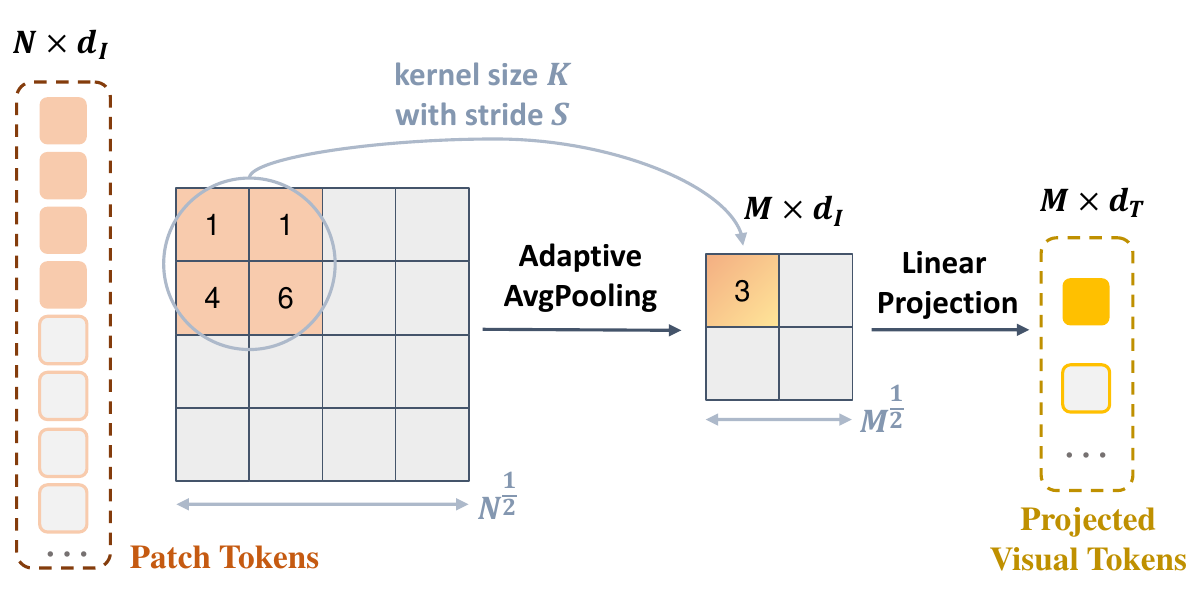} 
    \vspace{-10pt}
    \caption{\small Visualization of \model method.}
    \label{fig:method_avgpool}
    \vspace{-10pt}
\end{wrapfigure}

Inspired by the analysis in \textsection\ref{sec:analysis}, we propose a \model method to \textbf{De}couple vision token \textbf{Co}mpression from semantic abstraction in MLLMs.  In this approach, the compressive projectors focus on reducing the number of visual tokens with patch-level outcomes, while the LLM serves as the expert semantic abstractor. Consequently, the \model system only requires a simple projector that compresses visual tokens at the patch level. This design removes the intermediate semantic bottleneck and simplifies the training process.

We employ the 2D Adaptive Average Pooling (abbreviated as \avgpool) as a natural downsampler of the visual tokens at the patch level. As Figure~\ref{fig:method_avgpool} illustrates, given $N$ patch tokens from the ViT, the adaptive pooling can reduce the token number to a lesser square number $M$. Specifically, we reshape the N visual tokens to 2D tensors with size $(N^\frac{1}{2}, N^\frac{1}{2})$ and utilize a 2D adaptive average pooling to get compressed tokens with size  $(M^\frac{1}{2}, M^\frac{1}{2})$. Subsequently, the compressed 2D tensor is flattened into $M$ tokens. These tokens are finally projected by the linear layer to match the textual embedding dimension, serving as visual inputs to the LLM. During compression, the adaptive pooling~\footnote{ Apply the torch.nn.AdaptiveAvgPool2d function in the PyTorch framework.} automatically calculates the stride $S$ and kernel size $K$ in a parameter-free mode. It averages patches in a spatial $K\times K$ window into a mixed token. In essence, the 2D \avgpool merges the spatial neighbor patch tokens which tend to have high visual redundancy. 

As illustrated in the third row of Figure~\ref{fig:model_r-gae}, the Query-to-Patch mapping of the \avgpool projector forms a 2D grating pattern. It uniformly down-samples the grouped patches over the 2D spatial space of the original image. This uniform patch-level sampling preserves dense visual context compared to the QFormer abstractor. For instance, the compressed token indexed at $(3,3)$, highlighted in the red frame, retains the fine-grained representation of the ``purple and red buttons''. Subsequently, the LLM can attend to the accurate visual region by leveraging the visual context from the \avgpool, as shown in the Text-to-Patch map. Furthermore, the Text-to-Patch maps of the linear projector and AdaptiveAvgPool are nearly identical. This similarity reveals that the AdaptiveAvgPool projector achieves a superior combination of (i) effectiveness, approximating the linear projector in preserving visual context, and (ii) efficiency, reducing the number of vision tokens, similar to the QFormer abstractor.

\begin{table}[t]
\caption{
\small
Overall performance compared to existing compressive projectors. All results are conducted under the same architecture and settings: CLIP ViT-L/14~\cite{radford2021clip} as the vision backbone with $336^2$ image resolution and Vicuna-v1.5-7B~\cite{vicuna2023} as the LLM with LoRA~\cite{hu2021lora} tuning strategy. The training data contains 558K image-text pairs for pre-training and 665K chat instances for instruction tuning, the same as the LLaVA-1.5~\cite{liu2023llava15}. All compressive projectors reduce the vision token number (\#V) from 576 to 144. * indicates reproduced results using LoRA while ${\dagger}$ denotes the full-training results reported in LLaVA v1.5. The best and second-best results are \textbf{bolded} and \underline{underlined}, respectively.}
\label{tab:main_results}
\centering
\begin{adjustbox}{max width=\columnwidth}

\begin{tabular}{l ccc cccccccc}
\toprule
\multicolumn{1}{l}{\textbf{Projectors}} &
  \textbf{\#V} &
  \textbf{\seedimg} &
  \textbf{\mmep} &
  \multicolumn{1}{c}{\textbf{POPE}} &
  \multicolumn{1}{c}{\textbf{Refcoco}} &
  \multicolumn{1}{c}{\textbf{Refcoco+}} &
  \multicolumn{1}{c}{\textbf{Refcocog}} &
  \multicolumn{1}{c}{\textbf{VizWiz}} &
  \multicolumn{1}{c}{\textbf{\vqa}} &
  \multicolumn{1}{c}{\textbf{GQA}} &
  \multicolumn{1}{c}{\textbf{\textvqa}} \\ \midrule
\demph{$\text{Linear}^{\dagger}$}~\cite{liu2023llava15}           & \demph{576} & \demph{66.2} & \demph{1524.6} & \demph{86.4} & \demph{54.4} & \demph{47.8} & \demph{49.8} & \demph{53.6} & \demph{76.3} & \demph{60.0} & \demph{58.9} \\
Linear\textsuperscript{*}~\cite{liu2023llava15}           & 576 & 65.1 & 1338.6 & 86.8 & 46.9 & 41.6 & 46.3 & 50.2 & 74.9 & 56.5 & 58.4 \\ \midrule
QFormer~\cite{li2023blip2}              & 144 & 55.3 & 1312.7  & 79.0 & 15.1 & 10.5 & 11.6 & \textbf{51.2} & 65.6 & 48.6 & 50.7 \\
C-Abstractor~\cite{cha2023honeybee}     & 144 & \underline{60.5} & \textbf{1411.8}  & 84.5 & \underline{40.6} & \underline{34.3} & \underline{38.4} & 47.8 & 70.9 & 52.6 & \underline{55.9} \\
D-Abstractor~\cite{cha2023honeybee}     & 144 & 60.0 & 1313.2  & \underline{84.6} & 32.9 & 27.6 & 32.4 & \underline{49.7} & \underline{71.1} & \underline{53.1} & 55.1 \\
\model (Ours)                             & 144 & \textbf{62.8} & \underline{1373.4}  & \textbf{85.9} & \textbf{43.4} & \textbf{38.5} & \textbf{39.3} & \underline{49.7} & \textbf{74.0} & \textbf{54.1} & \textbf{56.2} \\ \bottomrule
\end{tabular}
\end{adjustbox}
\vspace{-20pt}
\end{table}

\begin{figure}[t!]
    \centering
    \begin{minipage}[b]{0.57\textwidth}
\begin{table}[H]
  \caption{
    \small
    Vision spatial understanding capability across different projectors.
    Task names are abbreviated as follows: Position (POS) for MME~\cite{fu2023mme}, Spatial Relationship (SR), Object Localization (OL), and Physical Relation (PR) for MMBench~\cite{ge2024mllmbench}, and Spatial Relation (SR) and Instance Location (IL) for SEED-Bench~\cite{li2023seedbench}.}
    \label{tab:main_locality}
    
    \centering
    \begin{adjustbox}{max width=\columnwidth}
    \begin{tabular}{l c cccccc c}
    \toprule
    \multirow{2}{*}{Projector} &
  \multirow{2}{*}{\#V} &
  \multicolumn{1}{c}{MME} &
  \multicolumn{3}{c}{MMB} &
  \multicolumn{2}{c}{SEED} &
  \multirow{2}{*}{Avg}  \\
  \addlinespace[2pt] \cmidrule(lr){3-3} \cmidrule(lr){4-6} \cmidrule(lr){7-8} \addlinespace
 &
   &
  \multicolumn{1}{l}{POS} &
  \multicolumn{1}{l}{SR} &
  \multicolumn{1}{l}{OL} &
  \multicolumn{1}{l}{PR} &
  \multicolumn{1}{l}{SR} &
  \multicolumn{1}{l}{IL} &   \\ \midrule
\demph{Linear~\cite{liu2023llava15}}       & \demph{576} & \demph{123.3}          & \demph{20.0}          & \demph{51.9}          & \demph{33.3}          & \demph{50.2}          & \demph{59.6}  &  \demph{56.4}       \\ \midrule
QFormer~\cite{li2023blip2}      & 144 & 73.3           & 17.8          & 33.3          & 33.3          & 39.0         & 48.9    &  40.9   \\
C-Abstractor~\cite{cha2023honeybee} & 144 & 116.7          & 15.6          & 42.0          & \textbf{54.2} & 43.5          & 54.4      & 54.4   \\
\model (Ours)         & 144 & \textbf{116.7} & \textbf{24.4} & \textbf{48.1} & 41.7          & \textbf{46.6} & \textbf{58.5} & \textbf{56.0} \\ \bottomrule
\end{tabular}
\end{adjustbox}
\end{table}
\end{minipage}
\hfill
\begin{minipage}[b]{0.4\textwidth}
  \centering
  \includegraphics[width=\textwidth]{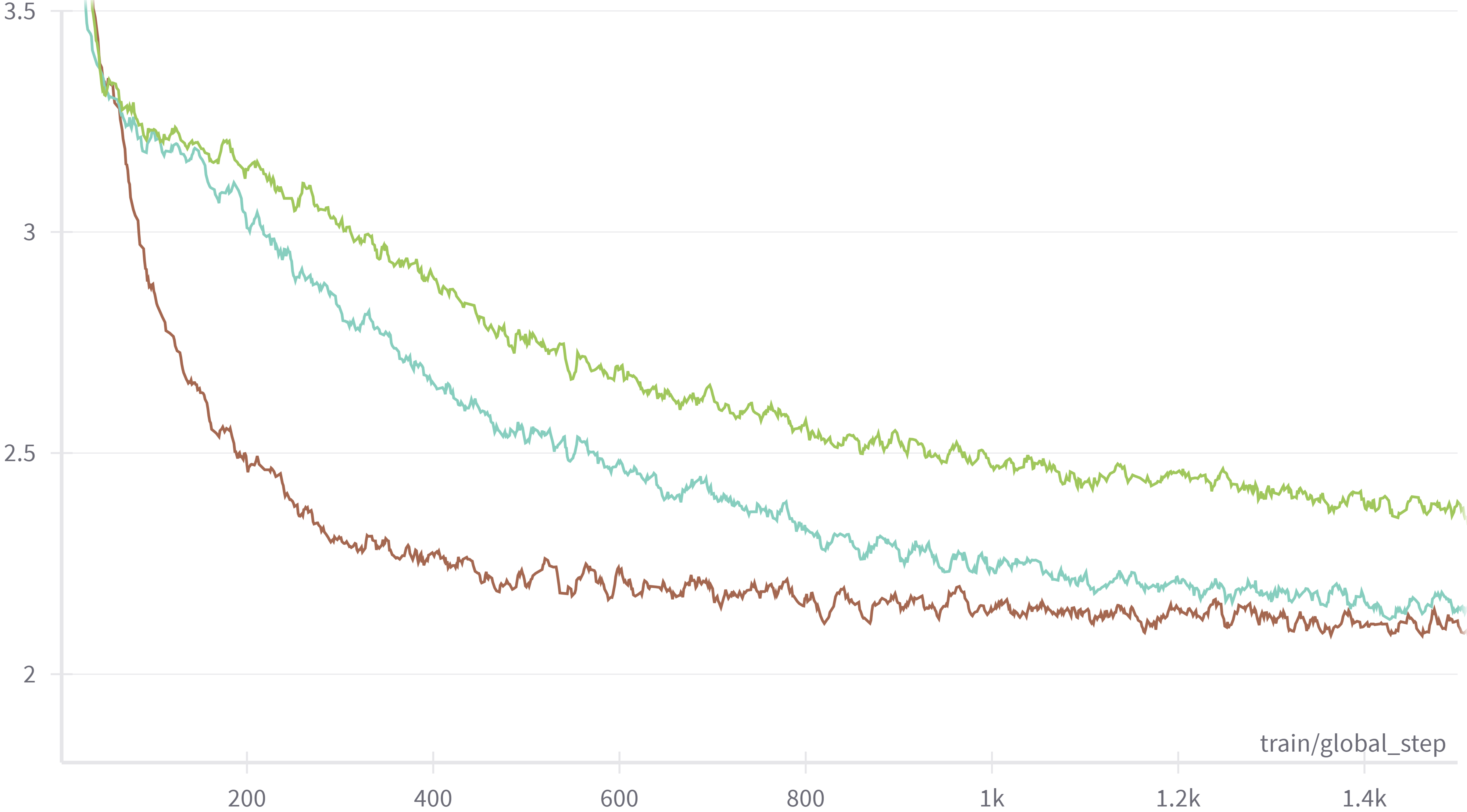}
  \caption{\small Pre-training loss convergence of \textcolor{brown}{\avgpool (brown)}, \textcolor{babyblue}{C-Abstractor (blue)} and \textcolor{ggreen}{QFormer (green)}.}
    \label{fig:exp_train-loss}
\end{minipage}
\end{figure}

\section{Experiments}
\subsection{Experiment Setting}

\paragraph{Training data and Evaluation.}
\label{subsec:training-data-eval}
We utilize the open-sourced 558K pre-training data (sourced from LAION~\cite{laion400m}, Conceptual Captions~\cite{Changpinyo2021Conceptual1P} and SBU Captions~\cite{ordonez2011sbu}) and 665K instruction-following data (containing LLaVa Synthetic Data~\cite{liu2023llava}, \vqa~\cite{balanced_vqa_v2}, GQA~\cite{hudson2019gqa}, OK-VQA~\cite{marino2019okvqa}, OCR-VQA~\cite{mishra2019ocr_vqa}, A-OKVQA~\cite{schwenk2022aokvqa}, TextCaps~\cite{sidorov2020textcaps}, 
 RefCOCO~\cite{yu2016modeling_refcoco}, Visual Genome~\cite{krishna2017visual_genome} and ShareGPT~\cite{ShareGPT2023}) following LLaVA v1.5~\cite{liu2023llava15}.
For evaluation, we measure model performance spanning three aspects. \textit{Multimodal LLM Benchmarks} including SEED-Bench~\cite{li2023seedbench} (report image-only set as \seedimg), MME~\cite{fu2023mme} (report perception set as \mmep) and POPE~\cite{li2023pope} are specially designed for instruction-following MLLMs. \textit{Visual Localization} task encompassing RefCOCO, RefCOCO+, and RefCOCOg~\cite{kazemzadeh2014referitgame, yu2016modeling_refcoco} is to measure the bounding box prediction accuracy. \textit{Open-Ended Visual Question Answering} task consisting of VizWiz~\cite{vizwiz}, \vqa~\cite{balanced_vqa_v2}, GQA~\cite{hudson2019gqa} and TextVQA~\cite{singh2019text_vqa} aims to evaluate visual reasoning capability.

\paragraph{Implementation Details.}
\label{subsec:exp_impl}
\model is primarily built on the LLaVA v1.5 framework, encompassing model architectures, training data, and training strategies. We replace the original two-layer MLP projector with QFormer~\cite{li2023blip2}, C-Abstractor~\cite{cha2023honeybee}, D-Abstractor~\cite{cha2023honeybee} and \avgpool respectively for fair comparison. The default configuration includes a CLIP ViT-L/14 \@ 336px~\cite{radford2021clip} and Vicuna v1.5 7B~\cite{vicuna2023} with a two-stage training strategy. The first pre-training stage updates only 
 the projector while the second instruction-tuning stage optimizes both the projector and the LLM using LoRA~\cite{hu2021lora}. The main results are derived from this default configuration. Additionally, we conduct generalization experiments using a more lightweight setup that involves only the instruction tuning stage as outlined in PRISM~\cite{prismatic}. Specific training hyper-parameters are detailed in Appendix C.

\subsection{Compared with Existing Projectors}
To showcase the efficiency and effectiveness of the \model method, we compare it with common projectors including the Linear projector~\cite{liu2023llava}, QFormer~\cite{li2023blip2}, C-Abstractor~\cite{cha2023honeybee}, and D-Abstractor\cite{cha2023honeybee}.

\begin{table}[t]
\caption{
\small
Comprehensive comparison between the C-Abstractor (C-Abstr) and Adaptive Averaging Pooling (AvgPool) across various settings including different vision backbones, image resolutions and LLMs. All experiments are conducted on the one-stage instruction tuning (665K data) referring to PRISM~\cite{prismatic} to speed up training. \textit{Res.} denotes image resolution. \textit{Compress.} means the compression ratio of each projector from the raw visual token number to the projected vision token number. }
\label{tab:generalization_results}
\centering
\begin{adjustbox}{max width=\columnwidth}
\begin{tabular}{@{}c@{\quad} cccc lcccccc@{}}
\toprule
 &
  \textbf{ViT} &
  \textbf{LLM} 
   &
  \textbf{Res.} &
  \textbf{Compress.} &
  \textbf{Project.} &
 \multicolumn{1}{c}{\textbf{POPE}} &
  \multicolumn{1}{c}{\textbf{Refcoco / + / g}} &
  \multicolumn{1}{c}{\textbf{VizWiz}} &
  \multicolumn{1}{c}{\textbf{\vqa}} &
  \multicolumn{1}{c}{\textbf{GQA}} &
  \multicolumn{1}{c}{\textbf{\textvqa}}  \\ \midrule 
\multirow{2}{*}{B1}  &
  \multirow{2}{*}{SigLIP ViT-SO~\cite{zhai2023siglip}} &
  Phi-2~\cite{phi2} &
  \multirow{2}{*}{224} & 
  \multirow{2}{*}{256-\textgreater{}144} &
  C-Abstr &
  66.1 &
  11.5 /
  6.5 /
  8.4 &
  18.7 &
  47.8 &
  42.0 &
  34.6 \\ \addlinespace
 &
   &(2.7B)
   & 
   &
   &
  AvgPool &
  \textbf{84.1} &
  \textbf{21.5} /
  \textbf{13.6} /
  \textbf{15.6} &
  \textbf{34.5} &
  \textbf{68.1} &
  \textbf{52.6} &
  \textbf{41.2} \\ \midrule \midrule
\multirow{2}{*}{B2} &
  
  \multirow{2}{*}{CLIP ViT-L~\cite{radford2021clip}} &
  Phi-2\ &
  \multirow{2}{*}{336} &
  \multirow{2}{*}{576-\textgreater{}144} &
  C-Abstr &
  73.7 &
  11.8 /
  7.3 /
  6.9 &
  18.0 &
  52.5 &
  45.3 &
  36.7 \\ \addlinespace
 &
   &(2.7B)
   &
   & 
   &
  AvgPool &
  \textbf{84.5} &
  \textbf{15.0} /
  \textbf{9.3} /
  \textbf{8.8} &
  \textbf{28.4} &
  \textbf{64.6} &
  \textbf{48.9} &
  \textbf{40.8} \\ \midrule \midrule
\multirow{2}{*}{B3} &
 
  \multirow{2}{*}{SigLIP ViT-SO} &
  Phi-2 &
   \multirow{2}{*}{384} &
  \multirow{2}{*}{729-\textgreater{}144} &
  C-Abstr &
   78.8&
   12.9 / 
   8.2 /
   7.7&
   \textbf{41.3}&
   53.2&
   45.1&
   35.4 \\ \addlinespace
 &
   &(2.7B)
   & 
   &
   &
  AvgPool &
   \textbf{81.7}&
   \textbf{17.4} /
   \textbf{11.4} / 
   \textbf{11.0} &
   39.5 &
   \textbf{60.3} &
   \textbf{48.0} &
   \textbf{40.2} \\ \midrule \midrule
\multirow{2}{*}{B4} &
  
  \multirow{2}{*}{DINOv2~\cite{oquab2023dinov2}+SigLIP} & Phi-2
   & 
   \multirow{2}{*}{384} &
  \multirow{2}{*}{729-\textgreater{}144} &
  C-Abstr &
  52.6 &
  13.5 /
  6.6 /
  7.5 &
  \textbf{29.2} &
  40.9 &
  36.3 &
  34.9 \\ \addlinespace
 &
   &(2.7B)
   & 
   &
   &
  AvgPool &
  \textbf{85.7} &
  \textbf{24.9} /
  \textbf{17.3} /
  \textbf{21.6} &
  24.0 &
  \textbf{63.9} &
  \textbf{52.6} &
  \textbf{39.2} \\ \midrule \midrule
\multirow{2}{*}{B5} &
 
  \multirow{2}{*}{DINOv2+SigLIP} &
  Qwen-Chat~\cite{bai2023qwen} &
   \multirow{2}{*}{384} &
  \multirow{2}{*}{729-\textgreater{}144} &
  C-Abstr &
  \textbf{49.9} &
  8.7 /
  4.3 /
  7.6 &
  17.7 &
  53.8 &
  45.1 &
  28.9 \\ \addlinespace
 &
   &(0.5B)
   & 
   &
   &
  AvgPool &
  \textbf{49.9} &
  \textbf{12.9} /
  \textbf{9.7} /
  \textbf{11.2} &
  \textbf{25.3} &
  \textbf{58.3} &
  \textbf{46.5} &
  \textbf{31.4} \\ \midrule \midrule
\multirow{2}{*}{B6} &

  \multirow{2}{*}{DINOv2+SigLIP} &
  Vicuna-v1.5~\cite{vicuna2023} &
    \multirow{2}{*}{384} &
  \multirow{2}{*}{729-\textgreater{}144} &
  C-Abstr &
   86.0 &
   31.7 /
   25.5 /
   29.2 &
   39.1 &
   62.6 &
   52.3 &
   46.5 \\ \addlinespace
 &
   & (7B)
   &
   &
   &
  AvgPool &
   \textbf{87.0} &
   \textbf{42.3} /
   \textbf{33.1} /
   \textbf{37.6} &
   \textbf{52.2} &
   \textbf{69.8} &
   \textbf{55.4} &
   \textbf{49.3} \\ \bottomrule
\end{tabular}
\end{adjustbox}
\end{table}

\vspace{-5pt}
\paragraph{Performance Effectiveness.}
Table~\ref{tab:main_results} presents the overall performance of different projectors. The non-compressive linear projector preserves all vision information and achieves the best overall performance. In the compressive projector category, \model outperforms existing solutions across most benchmarks. Specifically, \model achieves gain margins of \seedimg +2.3 and POPE +1.3 in the instruction-following MLLM benchmarks, RefCOCO/RefCOCO+/RefCOCOg +2.8/4.2/0.9 for visual localization, and \vqa +3.9, GQA +1.0, \textvqa +0.3 for open-ended visual question answering. The superior results of \model under the same compression ratio (576->144) demonstrate that naive compression at the patch level effectively transmits visual context while reducing the token number. Among the existing projectors, the locality-enhanced C-Abstractor produces results comparable to \model. Additionally, we observe that QFormer performs poorly on the visual localization task, particularly in predicting visual coordinates. This poor performance is due to the loss of spatial locality during projector compression, resulting in cumulative spatial context deficiency.


\paragraph{Training Efficiency.}
Besides the remarkable performance, \model also has efficiency advantages because it conducts parameter-free compression clarified in \textsection~\ref{sec:model}.
Among existing compressive projectors, the sub-optimal C-Abstractor comprises 3-layer ResNet blocks~\cite{xie2017resnet}, the adaptive average pooling and another 3-layer ResNet blocks. Meanwhile, we adopt a two-layer QFormer consisting of a self-attention and a cross-attention layer initialized from the BLIP-2~\cite{li2023blip2} pretraining weights.
Compared with them, the \avgpool in  \model method is more lightweight and efficient. Figure~\ref{fig:exp_train-loss} depicts that \model has faster training convergence during pre-training.

\paragraph{Spatial Locality Reservation.} Spatial understanding capability in vision modality is essential to achieve accurate visual location, fine-grained vision reasoning, object relation perception and etc.  We verify the spatial understanding capability of \model in Table~\ref{tab:main_locality} across six spatial understanding tasks from MLLM benchmarks. As Honeybee~\cite{cha2023honeybee} points out, the vanilla resampler architecture like QFormer will lose the visual spatial locality, therefore, it obtains a low average score of 40.9. The locality-enhanced projector, i.e., C-Abstractor, has remarkable improvements and achieves 54.4. Overall, the \model 
 with \avgpool well reserves the significant spatial context and achieves the closest score (56.0) to the linear projector (56.4). This benefits from the kernel and stride operation of  2D \avgpool similar to the convolutional network~\cite{li2021cnnsurvey}.

\begin{figure}[t]
    \centering
    \begin{minipage}{0.35\textwidth}
    \centering
    \includegraphics[width=\linewidth]{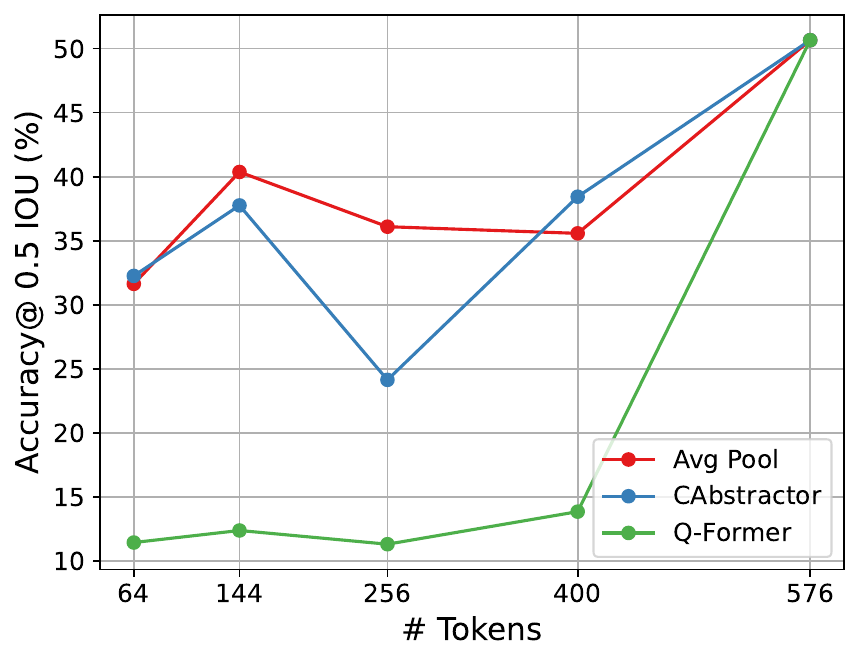} 
        \caption{\small Compression ratio reducing 576 tokens to 400/256/144/64 tokens.}
        \label{fig:abla_compressratio}
    \end{minipage}
    \hfill 
    \begin{minipage}{0.31\textwidth}
        \centering
        \includegraphics[width=\linewidth]{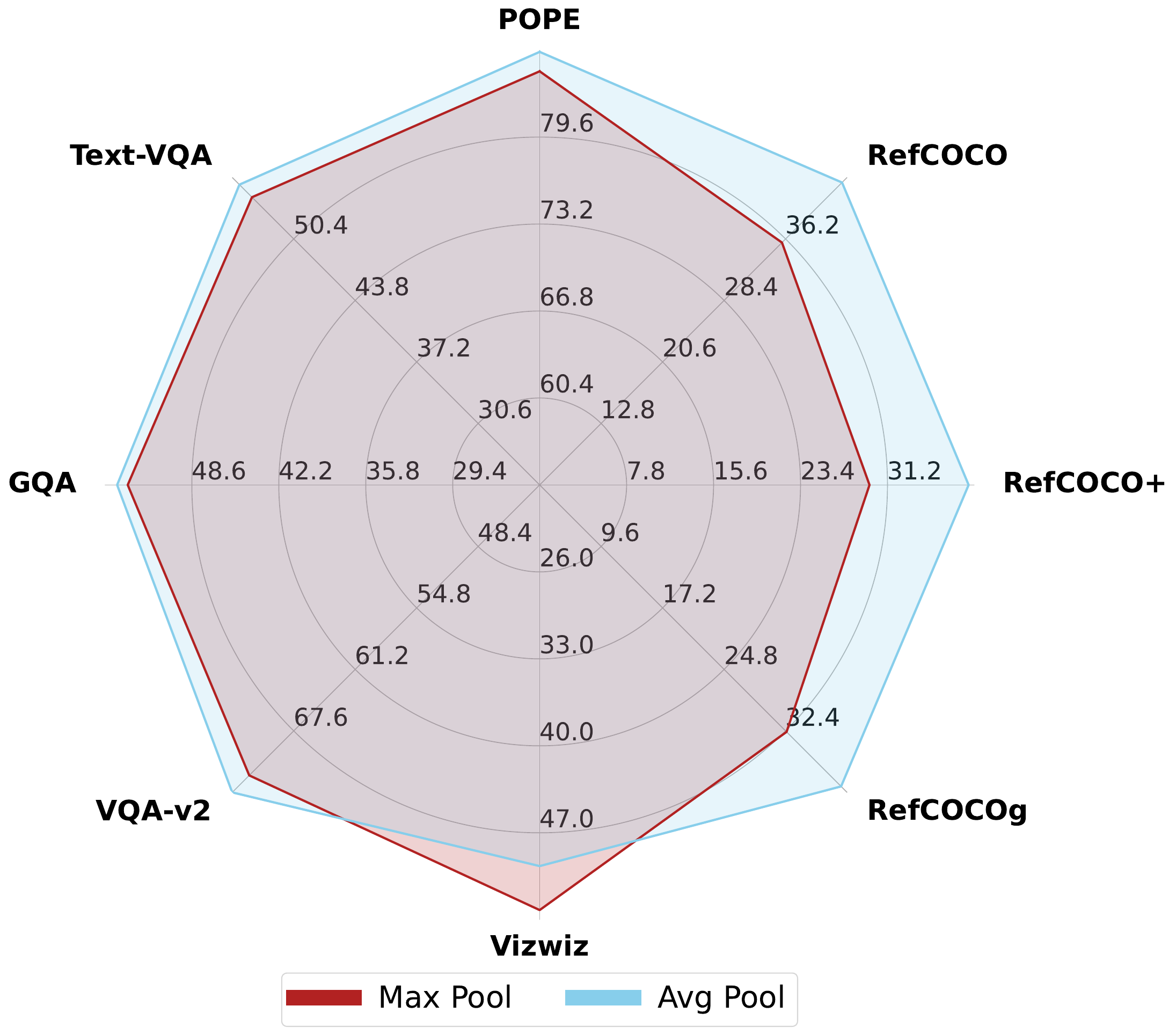} 
        \caption{
        \small 
        Comparison of Max Pooling and Average Pooling.}
        \label{fig:abla_maxpool}
    \end{minipage}
    \hfill 
    \begin{minipage}{0.31\textwidth}
    \centering
        \includegraphics[width=\linewidth]{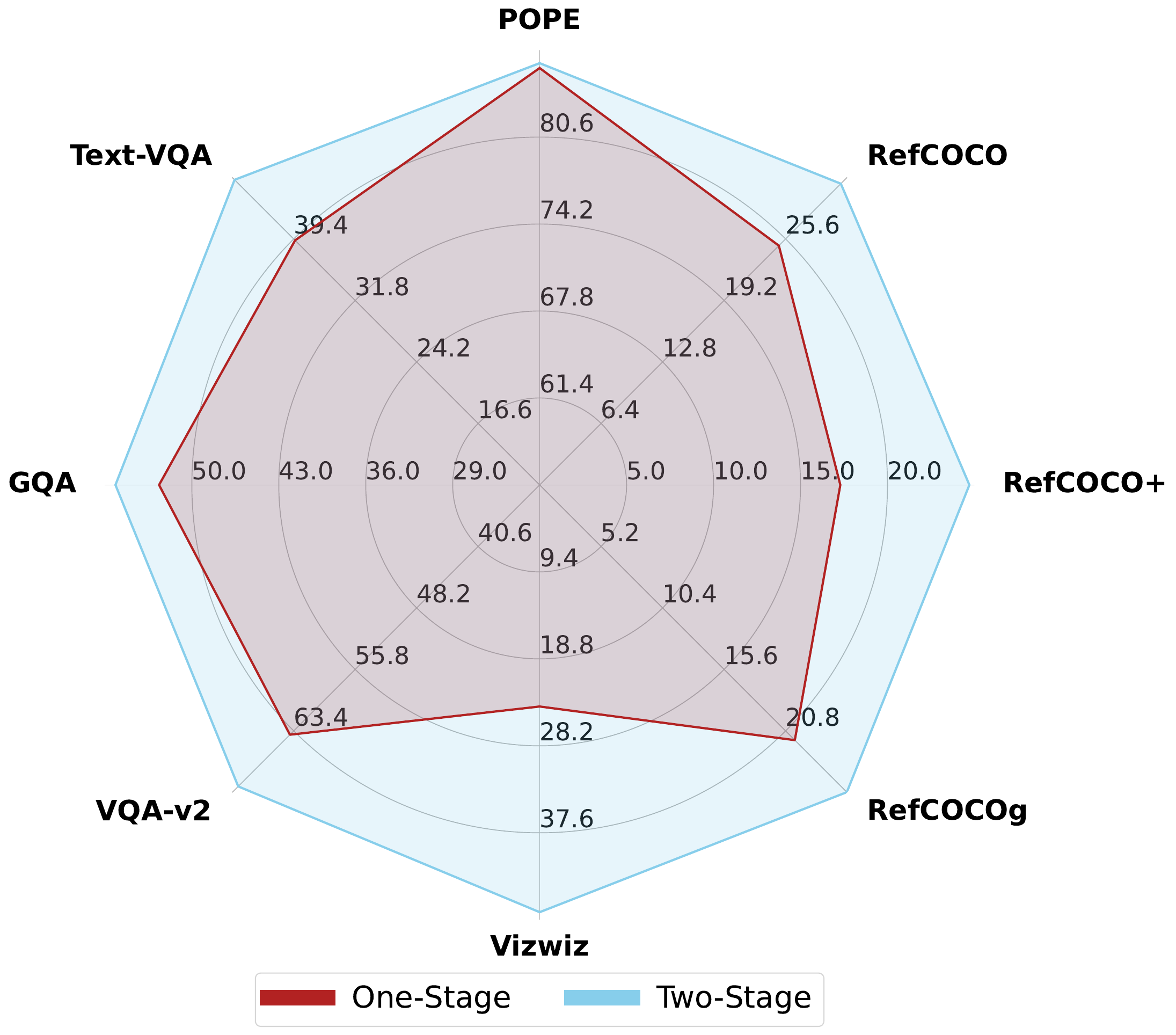} 
        \caption{
        \small 
        Comparison of one-stage and two-stage training.}
        \label{fig:abla_onestage}
    \end{minipage}
    \vspace{-10pt}
\end{figure}

\subsection{Generalization Results}

To explore the performance of \model under different configurations, we select varied vision backbones, image resolutions and LLMs, and report results in Table~\ref{tab:generalization_results}. To speed up training, all results are obtained through the one-stage training (i.e., instruction tuning) according to PRISM~\cite{prismatic}. We select the most comparative baseline C-Abstractor (refer to Table~\ref{tab:main_results}) as a reference.

For vision backbones (B2, B3, and B4), we adopt the CLIP ViT-L, SigLIP ViT-SO~\cite{zhai2023siglip}, and the DINOv2~\cite{oquab2023dinov2}+SigLIP ensemble in embedding dimension. For scaling image resolution (B1 and B3), we compare 224px and 384px image inputs using the SigLIP ViT-SO backbone. For LLMs (B4, B5, and B6), we employ three levels of model scope, including Qwen-Chat-0.5B~\cite{bai2023qwen}, Phi-2-2.7B~\cite{phi2}, and Vicuna-v1.5~\cite{vicuna2023}.

The overall results in Table~\ref{tab:generalization_results} under six different settings demonstrate the robustness of \model  as a compressive projector across diverse MLLM architectures. It surpasses the C-Abstractor notably in almost all metrics and all settings.

\subsection{Ablation Study}
\textbf{Compression Ratio Analysis.}
There is a trade-off between visual information deficiency and training cost based on the compression ratio. In Figure~\ref{fig:abla_compressratio}, we compress the visual tokens from $24\times24$ to $20\times20$, $16\times16$, $12\times12$, and $8\times8$ respectively, and report the average Accuracy@IoU=0.5 on the visual localization task. Results reveal that a quarter compression from  $24\times24$ to $12\times12$ provides the best balance for \avgpool.

\textbf{Average Pooling vs. Max Pooling.}
Average pooling and max pooling are two widely-used downsampling operations. We compare these two operations in the \model method in Figure~\ref{fig:abla_maxpool}. Results show that adaptive average pooling performs better across almost all metrics, especially visual localization. The reason is that the averaging operation integrates each patch within the kernel-size window and can serve more visual context.

\textbf{One-Stage vs. Multi-Stage Training.}
PRISM~\cite{prismatic} indicates simple linear projectors only require one-stage instruction tuning. Inspired by this, we compare the one-stage and two-stage training results of \model and find that two-stage training is recommended, as shown in Figure~\ref{fig:abla_onestage}.

\section{Limitations}
\label{sec:limitations}
We present limitations in this work to facilitate future research. 
Firstly, the \avgpool adopted in the \model method may cause severe visual information deficiency in an increasingly high compression ratio compared to semantic-level compression projectors. In a high-compression scenario, the averaging pooling will erase the fine-grained visual context in a kernel scope.
Secondly, the superiority of \model lies in a limited training resource application including limited GPUs to train a long visual token sequence and limited training data to optimize a desirable semantic QFormer-type projector. Otherwise, when have abundant training resources, the architecture of projectors tend to be insignificant in an MLLM system as pointed out in the MM1~\cite{mckinzie2024mm1}.

\section{Conclusion}
We introduce \model to decouple visual token compression from semantic abstraction. It is motivated by the ``Double Abstraction'' problem of existing projectors disentangling the Text-to-Patch, Text-to-Query and Query-to-Patch R-GAE maps in the vision-and-language semantic alignment. The \model method simplifies existing compressive projectors with a naive \avgpool, which downsamples spatial vision tokens directly at the spatial level. Experiments across diverse configurations demonstrate the efficiency,  effectiveness, and robustness of \model. Eventually, the intuition of ``DeCo'' is not limited to the specific \avgpool projector design, there is great potential to improve it to perform more effectively under more demanding scenarios like high compression ratio.

\bibliographystyle{abbrvnat}
\bibliography{neurips_2024}

\begin{thebibliography}{85}
\providecommand{\natexlab}[1]{#1}
\providecommand{\url}[1]{\texttt{#1}}
\expandafter\ifx\csname urlstyle\endcsname\relax
  \providecommand{\doi}[1]{doi: #1}\else
  \providecommand{\doi}{doi: \begingroup \urlstyle{rm}\Url}\fi

\bibitem[Abnar and Zuidema(2020)]{abnar2020quantifying}
S.~Abnar and W.~Zuidema.
\newblock Quantifying attention flow in transformers.
\newblock \emph{arXiv preprint arXiv:2005.00928}, 2020.

\bibitem[Aflalo et~al.(2022)Aflalo, Du, Tseng, Liu, Wu, Duan, and Lal]{aflalo2022vl}
E.~Aflalo, M.~Du, S.-Y. Tseng, Y.~Liu, C.~Wu, N.~Duan, and V.~Lal.
\newblock Vl-interpret: An interactive visualization tool for interpreting vision-language transformers.
\newblock In \emph{Proceedings of the IEEE/CVF Conference on computer vision and pattern recognition}, pages 21406--21415, 2022.

\bibitem[Alayrac et~al.(2022)Alayrac, Donahue, Luc, Miech, Barr, Hasson, Lenc, Mensch, Millican, Reynolds, Ring, Rutherford, Cabi, Han, Gong, Samangooei, Monteiro, Menick, Borgeaud, Brock, Nematzadeh, Sharifzadeh, Binkowski, Barreira, Vinyals, Zisserman, and Simonyan]{Alayrac2022FlamingoAV}
J.-B. Alayrac, J.~Donahue, P.~Luc, A.~Miech, I.~Barr, Y.~Hasson, K.~Lenc, A.~Mensch, K.~Millican, M.~Reynolds, R.~Ring, E.~Rutherford, S.~Cabi, T.~Han, Z.~Gong, S.~Samangooei, M.~Monteiro, J.~Menick, S.~Borgeaud, A.~Brock, A.~Nematzadeh, S.~Sharifzadeh, M.~Binkowski, R.~Barreira, O.~Vinyals, A.~Zisserman, and K.~Simonyan.
\newblock Flamingo: a visual language model for few-shot learning.
\newblock \emph{ArXiv}, abs/2204.14198, 2022.

\bibitem[Awadalla et~al.(2023)Awadalla, Gao, Gardner, Hessel, Hanafy, Zhu, Marathe, Bitton, Gadre, Sagawa, Jitsev, Kornblith, Koh, Ilharco, Wortsman, and Schmidt]{awadalla2023openflamingo}
A.~Awadalla, I.~Gao, J.~Gardner, J.~Hessel, Y.~Hanafy, W.~Zhu, K.~Marathe, Y.~Bitton, S.~Gadre, S.~Sagawa, J.~Jitsev, S.~Kornblith, P.~W. Koh, G.~Ilharco, M.~Wortsman, and L.~Schmidt.
\newblock Openflamingo: An open-source framework for training large autoregressive vision-language models.
\newblock \emph{ArXiv preprint}, abs/2308.01390, 2023.

\bibitem[Bai et~al.(2023{\natexlab{a}})Bai, Bai, Chu, Cui, Dang, Deng, Fan, Ge, Han, Huang, et~al.]{bai2023qwen}
J.~Bai, S.~Bai, Y.~Chu, Z.~Cui, K.~Dang, X.~Deng, Y.~Fan, W.~Ge, Y.~Han, F.~Huang, et~al.
\newblock Qwen technical report.
\newblock \emph{arXiv preprint arXiv:2309.16609}, 2023{\natexlab{a}}.

\bibitem[Bai et~al.(2023{\natexlab{b}})Bai, Bai, Yang, Wang, Tan, Wang, Lin, Zhou, and Zhou]{Qwen-VL}
J.~Bai, S.~Bai, S.~Yang, S.~Wang, S.~Tan, P.~Wang, J.~Lin, C.~Zhou, and J.~Zhou.
\newblock Qwen-vl: A frontier large vision-language model with versatile abilities.
\newblock \emph{ArXiv preprint}, abs/2308.12966, 2023{\natexlab{b}}.

\bibitem[Bavishi et~al.(2023)Bavishi, Elsen, Hawthorne, Nye, Odena, Somani, and Ta\c{s}\i{}rlar]{fuyu-8b}
R.~Bavishi, E.~Elsen, C.~Hawthorne, M.~Nye, A.~Odena, A.~Somani, and S.~Ta\c{s}\i{}rlar.
\newblock Introducing our multimodal models, 2023.
\newblock URL \url{https://www.adept.ai/blog/fuyu-8b}.

\bibitem[Ben Melech~Stan et~al.(2024)Ben Melech~Stan, Yehezkel~Rohekar, Gurwicz, Olson, Bhiwandiwalla, Aflalo, Wu, Duan, Tseng, and Lal]{ben2024lvlm-interpret}
G.~Ben Melech~Stan, R.~Yehezkel~Rohekar, Y.~Gurwicz, M.~L. Olson, A.~Bhiwandiwalla, E.~Aflalo, C.~Wu, N.~Duan, S.-Y. Tseng, and V.~Lal.
\newblock Lvlm-intrepret: An interpretability tool for large vision-language models.
\newblock \emph{arXiv e-prints}, pages arXiv--2404, 2024.

\bibitem[Bigham et~al.(2010)Bigham, Jayant, Ji, Little, Miller, Miller, Miller, Tatarowicz, White, White, et~al.]{vizwiz}
J.~P. Bigham, C.~Jayant, H.~Ji, G.~Little, A.~Miller, R.~C. Miller, R.~Miller, A.~Tatarowicz, B.~White, S.~White, et~al.
\newblock Vizwiz: nearly real-time answers to visual questions.
\newblock In \emph{Proceedings of the 23nd annual ACM symposium on User interface software and technology}, pages 333--342, 2010.

\bibitem[Carion et~al.(2020)Carion, Massa, Synnaeve, Usunier, Kirillov, and Zagoruyko]{carion2020detr}
N.~Carion, F.~Massa, G.~Synnaeve, N.~Usunier, A.~Kirillov, and S.~Zagoruyko.
\newblock End-to-end object detection with transformers.
\newblock In \emph{European conference on computer vision}, pages 213--229. Springer, 2020.

\bibitem[Cha et~al.(2024)Cha, Kang, Mun, and Roh]{cha2023honeybee}
J.~Cha, W.~Kang, J.~Mun, and B.~Roh.
\newblock Honeybee: Locality-enhanced projector for multimodal llm.
\newblock In \emph{Proceedings of the IEEE/CVF Conference on Computer Vision and Pattern Recognition (CVPR)}, 2024.

\bibitem[Changpinyo et~al.(2021)Changpinyo, Sharma, Ding, and Soricut]{Changpinyo2021Conceptual1P}
S.~Changpinyo, P.~K. Sharma, N.~Ding, and R.~Soricut.
\newblock Conceptual 12m: Pushing web-scale image-text pre-training to recognize long-tail visual concepts.
\newblock \emph{2021 IEEE/CVF Conference on Computer Vision and Pattern Recognition (CVPR)}, pages 3557--3567, 2021.

\bibitem[Chefer et~al.(2021{\natexlab{a}})Chefer, Gur, and Wolf]{chefer2021gae}
H.~Chefer, S.~Gur, and L.~Wolf.
\newblock Generic attention-model explainability for interpreting bi-modal and encoder-decoder transformers.
\newblock In \emph{Proceedings of the IEEE/CVF International Conference on Computer Vision}, pages 397--406, 2021{\natexlab{a}}.

\bibitem[Chefer et~al.(2021{\natexlab{b}})Chefer, Gur, and Wolf]{chefer2021transexplain}
H.~Chefer, S.~Gur, and L.~Wolf.
\newblock Transformer interpretability beyond attention visualization.
\newblock In \emph{Proceedings of the IEEE/CVF conference on computer vision and pattern recognition}, pages 782--791, 2021{\natexlab{b}}.

\bibitem[Chen et~al.(2023{\natexlab{a}})Chen, Zhang, Zeng, Zhang, Zhu, and Zhao]{chen2023shikra}
K.~Chen, Z.~Zhang, W.~Zeng, R.~Zhang, F.~Zhu, and R.~Zhao.
\newblock Shikra: Unleashing multimodal llm's referential dialogue magic.
\newblock \emph{arXiv preprint arXiv:2306.15195}, 2023{\natexlab{a}}.

\bibitem[Chen et~al.(2023{\natexlab{b}})Chen, Yao, and Jin]{chen2023rethinking}
W.~Chen, L.~Yao, and Q.~Jin.
\newblock Rethinking benchmarks for cross-modal image-text retrieval.
\newblock In \emph{Proceedings of the 46th International ACM SIGIR Conference on Research and Development in Information Retrieval}, pages 1241--1251, 2023{\natexlab{b}}.

\bibitem[Chen et~al.(2023{\natexlab{c}})Chen, Wu, Wang, Su, Chen, Xing, Zhong, Zhang, Zhu, Lu, Li, Luo, Lu, Qiao, and Dai]{internVL}
Z.~Chen, J.~Wu, W.~Wang, W.~Su, G.~Chen, S.~Xing, M.~Zhong, Q.~Zhang, X.~Zhu, L.~Lu, B.~Li, P.~Luo, T.~Lu, Y.~Qiao, and J.~Dai.
\newblock Internvl: Scaling up vision foundation models and aligning for generic visual-linguistic tasks.
\newblock \emph{arXiv preprint arXiv:2312.14238}, 2023{\natexlab{c}}.

\bibitem[Chiang et~al.(2023)Chiang, Li, Lin, Sheng, Wu, Zhang, Zheng, Zhuang, Zhuang, Gonzalez, Stoica, and Xing]{vicuna2023}
W.-L. Chiang, Z.~Li, Z.~Lin, Y.~Sheng, Z.~Wu, H.~Zhang, L.~Zheng, S.~Zhuang, Y.~Zhuang, J.~E. Gonzalez, I.~Stoica, and E.~P. Xing.
\newblock Vicuna: An open-source chatbot impressing gpt-4 with 90\%* chatgpt quality, March 2023.
\newblock URL \url{https://lmsys.org/blog/2023-03-30-vicuna/}.

\bibitem[Dai et~al.(2023)Dai, Li, Li, Tiong, Zhao, Wang, Li, Fung, and Hoi]{dai2023instructblip}
W.~Dai, J.~Li, D.~Li, A.~M.~H. Tiong, J.~Zhao, W.~Wang, B.~Li, P.~Fung, and S.~Hoi.
\newblock Instructblip: Towards general-purpose vision-language models with instruction tuning.
\newblock \emph{ArXiv preprint}, abs/2305.06500, 2023.

\bibitem[Devlin et~al.(2019)Devlin, Chang, Lee, and Toutanova]{devlin2019bert}
J.~Devlin, M.-W. Chang, K.~Lee, and K.~Toutanova.
\newblock {BERT}: Pre-training of deep bidirectional transformers for language understanding.
\newblock In \emph{Proceedings of the 2019 Conference of the North {A}merican Chapter of the Association for Computational Linguistics: Human Language Technologies, Volume 1 (Long and Short Papers)}, pages 4171--4186, 2019.

\bibitem[Dong et~al.(2022)Dong, Li, Dai, Zheng, Wu, Chang, Sun, Xu, Li, and Sui]{icl_survey}
Q.~Dong, L.~Li, D.~Dai, C.~Zheng, Z.~Wu, B.~Chang, X.~Sun, J.~Xu, L.~Li, and Z.~Sui.
\newblock A survey for in-context learning, 2022.

\bibitem[Dosovitskiy et~al.(2020)Dosovitskiy, Beyer, Kolesnikov, Weissenborn, Zhai, Unterthiner, Dehghani, Minderer, Heigold, Gelly, et~al.]{dosovitskiy2020image}
A.~Dosovitskiy, L.~Beyer, A.~Kolesnikov, D.~Weissenborn, X.~Zhai, T.~Unterthiner, M.~Dehghani, M.~Minderer, G.~Heigold, S.~Gelly, et~al.
\newblock An image is worth 16x16 words: Transformers for image recognition at scale.
\newblock \emph{arXiv preprint arXiv:2010.11929}, 2020.

\bibitem[Fu et~al.(2023)Fu, Chen, Shen, Qin, Zhang, Lin, Qiu, Lin, Yang, Zheng, Li, Sun, and Ji]{fu2023mme}
C.~Fu, P.~Chen, Y.~Shen, Y.~Qin, M.~Zhang, X.~Lin, Z.~Qiu, W.~Lin, J.~Yang, X.~Zheng, K.~Li, X.~Sun, and R.~Ji.
\newblock Mme: A comprehensive evaluation benchmark for multimodal large language models.
\newblock \emph{arXiv preprint arXiv:2306.13394}, 2023.

\bibitem[Ge et~al.(2024)Ge, Chen, Chen, Chen, Chen, Yan, Zhu, Lin, Xie, Zhang, Chai, Liu, Song, Wang, Gao, Zhang, Li, Wan, and Wang]{ge2024mllmbench}
W.~Ge, S.~Chen, G.~H. Chen, Z.~Chen, J.~Chen, S.~Yan, C.~Zhu, Z.~Lin, W.~Xie, X.~Zhang, Y.~Chai, X.~Liu, D.~Song, X.~Wang, A.~Gao, Z.~Zhang, J.~Li, X.~Wan, and B.~Wang.
\newblock Mllm-bench: Evaluating multimodal llms with per-sample criteria, 2024.

\bibitem[{Gemini Team}(2023)]{gemini}
{Gemini Team}.
\newblock Gemini: a family of highly capable multimodal models.
\newblock \emph{arXiv preprint arXiv:2312.11805}, 2023.

\bibitem[Goyal et~al.(2017)Goyal, Khot, Summers{-}Stay, Batra, and Parikh]{balanced_vqa_v2}
Y.~Goyal, T.~Khot, D.~Summers{-}Stay, D.~Batra, and D.~Parikh.
\newblock Making the {V} in {VQA} matter: Elevating the role of image understanding in visual question answering.
\newblock In \emph{2017 {IEEE} Conference on Computer Vision and Pattern Recognition, {CVPR} 2017, Honolulu, HI, USA, July 21-26, 2017}, pages 6325--6334, 2017.

\bibitem[Hu et~al.(2022)Hu, Shen, Wallis, Allen{-}Zhu, Li, Wang, Wang, and Chen]{hu2021lora}
E.~J. Hu, Y.~Shen, P.~Wallis, Z.~Allen{-}Zhu, Y.~Li, S.~Wang, L.~Wang, and W.~Chen.
\newblock Lora: Low-rank adaptation of large language models.
\newblock In \emph{The Tenth International Conference on Learning Representations, {ICLR} 2022, Virtual Event, April 25-29, 2022}, 2022.

\bibitem[Hudson and Manning(2019)]{hudson2019gqa}
D.~A. Hudson and C.~D. Manning.
\newblock {GQA:} {A} new dataset for real-world visual reasoning and compositional question answering.
\newblock In \emph{{IEEE} Conference on Computer Vision and Pattern Recognition, {CVPR} 2019, Long Beach, CA, USA, June 16-20, 2019}, pages 6700--6709, 2019.

\bibitem[Javaheripi et~al.(2023)Javaheripi, Bubeck, Abdin, Aneja, Bubeck, Mendes, Chen, Del~Giorno, Eldan, Gopi, et~al.]{phi2}
M.~Javaheripi, S.~Bubeck, M.~Abdin, J.~Aneja, S.~Bubeck, C.~C.~T. Mendes, W.~Chen, A.~Del~Giorno, R.~Eldan, S.~Gopi, et~al.
\newblock Phi-2: The surprising power of small language models.
\newblock \emph{Microsoft Research Blog}, 2023.

\bibitem[Kafle et~al.(2018)Kafle, Price, Cohen, and Kanan]{kafle2018dvqa}
K.~Kafle, B.~Price, S.~Cohen, and C.~Kanan.
\newblock Dvqa: Understanding data visualizations via question answering.
\newblock In \emph{Proceedings of the IEEE conference on computer vision and pattern recognition}, pages 5648--5656, 2018.

\bibitem[Karamcheti et~al.(2024)Karamcheti, Nair, Balakrishna, Liang, Kollar, and Sadigh]{prismatic}
S.~Karamcheti, S.~Nair, A.~Balakrishna, P.~Liang, T.~Kollar, and D.~Sadigh.
\newblock Prismatic vlms: Investigating the design space of visually-conditioned language models.
\newblock \emph{arXiv preprint arXiv:2402.07865}, 2024.

\bibitem[Kazemzadeh et~al.(2014)Kazemzadeh, Ordonez, Matten, and Berg]{kazemzadeh2014referitgame}
S.~Kazemzadeh, V.~Ordonez, M.~Matten, and T.~Berg.
\newblock Referitgame: Referring to objects in photographs of natural scenes.
\newblock In \emph{Proceedings of the 2014 conference on empirical methods in natural language processing (EMNLP)}, pages 787--798, 2014.

\bibitem[Krishna et~al.(2017)Krishna, Zhu, Groth, Johnson, Hata, Kravitz, Chen, Kalantidis, Li, Shamma, et~al.]{krishna2017visual_genome}
R.~Krishna, Y.~Zhu, O.~Groth, J.~Johnson, K.~Hata, J.~Kravitz, S.~Chen, Y.~Kalantidis, L.-J. Li, D.~A. Shamma, et~al.
\newblock Visual genome: Connecting language and vision using crowdsourced dense image annotations.
\newblock \emph{International journal of computer vision}, 123:\penalty0 32--73, 2017.

\bibitem[Laurençon et~al.(2023)Laurençon, Saulnier, Tronchon, Bekman, Singh, Lozhkov, Wang, Karamcheti, Rush, Kiela, Cord, and Sanh]{laurencon2023obelics}
H.~Laurençon, L.~Saulnier, L.~Tronchon, S.~Bekman, A.~Singh, A.~Lozhkov, T.~Wang, S.~Karamcheti, A.~M. Rush, D.~Kiela, M.~Cord, and V.~Sanh.
\newblock Obelics: An open web-scale filtered dataset of interleaved image-text documents, 2023.

\bibitem[Li et~al.(2023{\natexlab{a}})Li, Wang, Wang, Ge, Ge, and Shan]{li2023seedbench}
B.~Li, R.~Wang, G.~Wang, Y.~Ge, Y.~Ge, and Y.~Shan.
\newblock Seed-bench: Benchmarking multimodal llms with generative comprehension.
\newblock \emph{arXiv preprint arXiv:2307.16125}, 2023{\natexlab{a}}.

\bibitem[Li et~al.(2023{\natexlab{b}})Li, Zhang, Yang, Zhang, Pu, and Liu]{li2023otterhd}
B.~Li, P.~Zhang, J.~Yang, Y.~Zhang, F.~Pu, and Z.~Liu.
\newblock Otterhd: A high-resolution multi-modality model.
\newblock \emph{arXiv preprint arXiv:2311.04219}, 2023{\natexlab{b}}.

\bibitem[Li et~al.(2023{\natexlab{c}})Li, Zhang, Chen, Wang, Yang, and Liu]{li2023otter}
B.~Li, Y.~Zhang, L.~Chen, J.~Wang, J.~Yang, and Z.~Liu.
\newblock Otter: A multi-modal model with in-context instruction tuning.
\newblock \emph{ArXiv preprint}, abs/2305.03726, 2023{\natexlab{c}}.

\bibitem[Li et~al.(2023{\natexlab{d}})Li, Li, Savarese, and Hoi]{li2023blip2}
J.~Li, D.~Li, S.~Savarese, and S.~Hoi.
\newblock Blip-2: Bootstrapping language-image pre-training with frozen image encoders and large language models.
\newblock \emph{ArXiv preprint}, abs/2301.12597, 2023{\natexlab{d}}.

\bibitem[Li et~al.(2023{\natexlab{e}})Li, Xie, Li, Chen, Wang, Chen, Yang, Wang, and Kong]{2023vlfeedback}
L.~Li, Z.~Xie, M.~Li, S.~Chen, P.~Wang, L.~Chen, Y.~Yang, B.~Wang, and L.~Kong.
\newblock Silkie: Preference distillation for large visual language models.
\newblock 2023{\natexlab{e}}.

\bibitem[Li et~al.(2023{\natexlab{f}})Li, Yin, Li, Chen, Wang, Ren, Li, Yang, Xu, Sun, Kong, and Liu]{li2023m3it}
L.~Li, Y.~Yin, S.~Li, L.~Chen, P.~Wang, S.~Ren, M.~Li, Y.~Yang, J.~Xu, X.~Sun, L.~Kong, and Q.~Liu.
\newblock {M$^3$IT}: A large-scale dataset towards multi-modal multilingual instruction tuning.
\newblock \emph{ArXiv preprint}, abs/2306.04387, 2023{\natexlab{f}}.

\bibitem[Li et~al.(2024)Li, Wang, Xu, Wang, Feng, Kong, and Liu]{li2024multimodal}
L.~Li, Y.~Wang, R.~Xu, P.~Wang, X.~Feng, L.~Kong, and Q.~Liu.
\newblock Multimodal arxiv: A dataset for improving scientific comprehension of large vision-language models, 2024.

\bibitem[Li et~al.(2023{\natexlab{g}})Li, Du, Zhou, Wang, Zhao, and Wen]{li2023pope}
Y.~Li, Y.~Du, K.~Zhou, J.~Wang, W.~X. Zhao, and J.-R. Wen.
\newblock Evaluating object hallucination in large vision-language models.
\newblock In \emph{Proceedings of the 2023 Conference on Empirical Methods in Natural Language Processing}, pages 292--305, 2023{\natexlab{g}}.

\bibitem[Li et~al.(2021)Li, Liu, Yang, Peng, and Zhou]{li2021cnnsurvey}
Z.~Li, F.~Liu, W.~Yang, S.~Peng, and J.~Zhou.
\newblock A survey of convolutional neural networks: analysis, applications, and prospects.
\newblock \emph{IEEE transactions on neural networks and learning systems}, 33\penalty0 (12):\penalty0 6999--7019, 2021.

\bibitem[Liu et~al.(2023{\natexlab{a}})Liu, Li, Li, and Lee]{liu2023llava15}
H.~Liu, C.~Li, Y.~Li, and Y.~J. Lee.
\newblock Improved baselines with visual instruction tuning, 2023{\natexlab{a}}.

\bibitem[Liu et~al.(2023{\natexlab{b}})Liu, Li, Wu, and Lee]{liu2023llava}
H.~Liu, C.~Li, Q.~Wu, and Y.~J. Lee.
\newblock Visual instruction tuning.
\newblock \emph{ArXiv preprint}, abs/2304.08485, 2023{\natexlab{b}}.

\bibitem[Liu et~al.(2024{\natexlab{a}})Liu, Li, Li, Li, Zhang, Shen, and Lee]{llavanext}
H.~Liu, C.~Li, Y.~Li, B.~Li, Y.~Zhang, S.~Shen, and Y.~J. Lee.
\newblock Llava-next: Improved reasoning, ocr, and world knowledge, January 2024{\natexlab{a}}.
\newblock URL \url{https://llava-vl.github.io/blog/2024-01-30-llava-next/}.

\bibitem[Liu et~al.(2024{\natexlab{b}})Liu, Yan, Zaharia, and Abbeel]{liu2024lwm}
H.~Liu, W.~Yan, M.~Zaharia, and P.~Abbeel.
\newblock World model on million-length video and language with blockwise ringattention, 2024{\natexlab{b}}.

\bibitem[Liu et~al.(2023{\natexlab{c}})Liu, Ma, Schubert, Ouyang, Rong, and Xiong]{liu2023multimodal}
Z.~Liu, Y.~Ma, M.~Schubert, Y.~Ouyang, W.~Rong, and Z.~Xiong.
\newblock Multimodal contrastive transformer for explainable recommendation.
\newblock \emph{IEEE Transactions on Computational Social Systems}, 2023{\natexlab{c}}.

\bibitem[Lyu et~al.(2022)Lyu, Liang, Deng, Salakhutdinov, and Morency]{lyu2022dime}
Y.~Lyu, P.~P. Liang, Z.~Deng, R.~Salakhutdinov, and L.-P. Morency.
\newblock Dime: Fine-grained interpretations of multimodal models via disentangled local explanations.
\newblock In \emph{Proceedings of the 2022 AAAI/ACM Conference on AI, Ethics, and Society}, pages 455--467, 2022.

\bibitem[Ma et~al.(2024)Ma, Chu, Yang, Lin, Gao, and Zhao]{ma2024paramtuning}
X.~Ma, X.~Chu, Z.~Yang, Y.~Lin, X.~Gao, and J.~Zhao.
\newblock Parameter efficient quasi-orthogonal fine-tuning via givens rotation.
\newblock \emph{arXiv preprint arXiv:2404.04316}, 2024.

\bibitem[Madureira(2021)]{madureira-2021-flamingos}
B.~Madureira.
\newblock Flamingos and hedgehogs in the croquet-ground: Teaching evaluation of {NLP} systems for undergraduate students.
\newblock In \emph{Proceedings of the Fifth Workshop on Teaching NLP}, pages 87--91, 2021.

\bibitem[Marino et~al.(2019)Marino, Rastegari, Farhadi, and Mottaghi]{marino2019okvqa}
K.~Marino, M.~Rastegari, A.~Farhadi, and R.~Mottaghi.
\newblock {OK-VQA:} {A} visual question answering benchmark requiring external knowledge.
\newblock In \emph{{IEEE} Conference on Computer Vision and Pattern Recognition, {CVPR} 2019, Long Beach, CA, USA, June 16-20, 2019}, pages 3195--3204, 2019.

\bibitem[McKinzie et~al.(2024)McKinzie, Gan, Fauconnier, Dodge, Zhang, Dufter, Shah, Du, Peng, Weers, et~al.]{mckinzie2024mm1}
B.~McKinzie, Z.~Gan, J.-P. Fauconnier, S.~Dodge, B.~Zhang, P.~Dufter, D.~Shah, X.~Du, F.~Peng, F.~Weers, et~al.
\newblock Mm1: Methods, analysis \& insights from multimodal llm pre-training.
\newblock \emph{arXiv preprint arXiv:2403.09611}, 2024.

\bibitem[Mishra et~al.(2019)Mishra, Shekhar, Singh, and Chakraborty]{mishra2019ocr_vqa}
A.~Mishra, S.~Shekhar, A.~K. Singh, and A.~Chakraborty.
\newblock Ocr-vqa: Visual question answering by reading text in images.
\newblock In \emph{2019 international conference on document analysis and recognition (ICDAR)}, pages 947--952. IEEE, 2019.

\bibitem[OpenAI(2023)]{gpt4v}
OpenAI.
\newblock Gpt-4v(ision) system card, 2023.

\bibitem[Oquab et~al.(2023)Oquab, Darcet, Moutakanni, Vo, Szafraniec, Khalidov, Fernandez, Haziza, Massa, El-Nouby, et~al.]{oquab2023dinov2}
M.~Oquab, T.~Darcet, T.~Moutakanni, H.~Vo, M.~Szafraniec, V.~Khalidov, P.~Fernandez, D.~Haziza, F.~Massa, A.~El-Nouby, et~al.
\newblock Dinov2: Learning robust visual features without supervision.
\newblock \emph{arXiv preprint arXiv:2304.07193}, 2023.

\bibitem[Ordonez et~al.(2011)Ordonez, Kulkarni, and Berg]{ordonez2011sbu}
V.~Ordonez, G.~Kulkarni, and T.~Berg.
\newblock Im2text: Describing images using 1 million captioned photographs.
\newblock \emph{Advances in neural information processing systems}, 24, 2011.

\bibitem[Radford et~al.(2021)Radford, Kim, Hallacy, Ramesh, Goh, Agarwal, Sastry, Askell, Mishkin, Clark, Krueger, and Sutskever]{radford2021clip}
A.~Radford, J.~W. Kim, C.~Hallacy, A.~Ramesh, G.~Goh, S.~Agarwal, G.~Sastry, A.~Askell, P.~Mishkin, J.~Clark, G.~Krueger, and I.~Sutskever.
\newblock Learning transferable visual models from natural language supervision.
\newblock In \emph{International Conference on Machine Learning}, 2021.

\bibitem[Ramesh and Koh(2022)]{ramesh2022investigation}
K.~Ramesh and Y.~S. Koh.
\newblock Investigation of explainability techniques for multimodal transformers.
\newblock In \emph{Australasian Conference on Data Mining}, pages 90--98. Springer, 2022.

\bibitem[Reka(2024)]{reka2024core}
T.~Reka.
\newblock Reka {C}ore, {F}lash, and {E}dge: A series of powerful multimodal language models, 2024.

\bibitem[Ren et~al.(2021)Ren, Lin, Zhao, Men, Yang, Zhou, Sun, and Yang]{ren2021iais}
S.~Ren, J.~Lin, G.~Zhao, R.~Men, A.~Yang, J.~Zhou, X.~Sun, and H.~Yang.
\newblock Learning relation alignment for calibrated cross-modal retrieval.
\newblock In \emph{Proceedings of the 59th Annual Meeting of the Association for Computational Linguistics and the 11th International Joint Conference on Natural Language Processing (Volume 1: Long Papers)}, 2021.

\bibitem[Ren et~al.(2023{\natexlab{a}})Ren, Chen, Li, Sun, and Hou]{testa}
S.~Ren, S.~Chen, S.~Li, X.~Sun, and L.~Hou.
\newblock {TESTA}: Temporal-spatial token aggregation for long-form video-language understanding.
\newblock In \emph{Findings of the Association for Computational Linguistics: EMNLP 2023}. Association for Computational Linguistics, Dec. 2023{\natexlab{a}}.

\bibitem[Ren et~al.(2023{\natexlab{b}})Ren, Li, Ren, Zhao, and Sun]{clip-openness}
S.~Ren, L.~Li, X.~Ren, G.~Zhao, and X.~Sun.
\newblock Delving into the openness of {CLIP}.
\newblock In \emph{Findings of the Association for Computational Linguistics: ACL 2023}. Association for Computational Linguistics, July 2023{\natexlab{b}}.

\bibitem[Ren et~al.(2023{\natexlab{c}})Ren, Yao, Li, Sun, and Hou]{timechat}
S.~Ren, L.~Yao, S.~Li, X.~Sun, and L.~Hou.
\newblock Timechat: A time-sensitive multimodal large language model for long video understanding.
\newblock \emph{ArXiv}, abs/2312.02051, 2023{\natexlab{c}}.

\bibitem[Ren et~al.(2024)Ren, Zhang, Zhu, Zhang, Zheng, Li, Smola, and Sun]{ren2024prompt}
S.~Ren, A.~Zhang, Y.~Zhu, S.~Zhang, S.~Zheng, M.~Li, A.~J. Smola, and X.~Sun.
\newblock Prompt pre-training with twenty-thousand classes for open-vocabulary visual recognition.
\newblock \emph{Advances in Neural Information Processing Systems}, 36, 2024.

\bibitem[Rohekar et~al.(2023)Rohekar, Gurwicz, and Nisimov]{rohekar2024causal}
R.~Y. Rohekar, Y.~Gurwicz, and S.~Nisimov.
\newblock Causal interpretation of self-attention in pre-trained transformers.
\newblock \emph{Advances in Neural Information Processing Systems}, 36, 2023.

\bibitem[Schuhmann et~al.(2021)Schuhmann, Vencu, Beaumont, Kaczmarczyk, Mullis, Katta, Coombes, Jitsev, and Komatsuzaki]{laion400m}
C.~Schuhmann, R.~Vencu, R.~Beaumont, R.~Kaczmarczyk, C.~Mullis, A.~Katta, T.~Coombes, J.~Jitsev, and A.~Komatsuzaki.
\newblock Laion-400m: Open dataset of clip-filtered 400 million image-text pairs.
\newblock \emph{ArXiv preprint}, abs/2111.02114, 2021.

\bibitem[Schwenk et~al.(2022)Schwenk, Khandelwal, Clark, Marino, and Mottaghi]{schwenk2022aokvqa}
D.~Schwenk, A.~Khandelwal, C.~Clark, K.~Marino, and R.~Mottaghi.
\newblock A-okvqa: A benchmark for visual question answering using world knowledge.
\newblock In \emph{Computer Vision--ECCV 2022: 17th European Conference, Tel Aviv, Israel, October 23--27, 2022, Proceedings, Part VIII}, pages 146--162. Springer, 2022.

\bibitem[ShareGPT(2023)]{ShareGPT2023}
ShareGPT.
\newblock Sharegpt.
\newblock 2023.

\bibitem[Sidorov et~al.(2020)Sidorov, Hu, Rohrbach, and Singh]{sidorov2020textcaps}
O.~Sidorov, R.~Hu, M.~Rohrbach, and A.~Singh.
\newblock Textcaps: a dataset for image captioning with reading comprehension.
\newblock In \emph{Computer Vision--ECCV 2020: 16th European Conference, Glasgow, UK, August 23--28, 2020, Proceedings, Part II 16}, pages 742--758. Springer, 2020.

\bibitem[Singh et~al.(2019)Singh, Natarajan, Shah, Jiang, Chen, Batra, Parikh, and Rohrbach]{singh2019text_vqa}
A.~Singh, V.~Natarajan, M.~Shah, Y.~Jiang, X.~Chen, D.~Batra, D.~Parikh, and M.~Rohrbach.
\newblock Towards {VQA} models that can read.
\newblock In \emph{{IEEE} Conference on Computer Vision and Pattern Recognition, {CVPR} 2019, Long Beach, CA, USA, June 16-20, 2019}, pages 8317--8326, 2019.

\bibitem[Song et~al.(2023)Song, Chai, Wang, Zhang, Zhou, Wu, Guo, Ye, Lu, Hwang, et~al.]{song2023moviechat}
E.~Song, W.~Chai, G.~Wang, Y.~Zhang, H.~Zhou, F.~Wu, X.~Guo, T.~Ye, Y.~Lu, J.-N. Hwang, et~al.
\newblock Moviechat: From dense token to sparse memory for long video understanding.
\newblock \emph{arXiv preprint arXiv:2307.16449}, 2023.

\bibitem[Swamy et~al.(2024)Swamy, Satayeva, Frej, Bossy, Vogels, Jaggi, K{\"a}ser, and Hartley]{swamy2024multimodn}
V.~Swamy, M.~Satayeva, J.~Frej, T.~Bossy, T.~Vogels, M.~Jaggi, T.~K{\"a}ser, and M.-A. Hartley.
\newblock Multimodn—multimodal, multi-task, interpretable modular networks.
\newblock \emph{Advances in Neural Information Processing Systems}, 36, 2024.

\bibitem[Vaswani et~al.(2017)Vaswani, Shazeer, Parmar, Uszkoreit, Jones, Gomez, Kaiser, and Polosukhin]{vaswani2017attention}
A.~Vaswani, N.~Shazeer, N.~Parmar, J.~Uszkoreit, L.~Jones, A.~N. Gomez, {\L}.~Kaiser, and I.~Polosukhin.
\newblock Attention is all you need.
\newblock \emph{Advances in neural information processing systems}, 30, 2017.

\bibitem[Voita et~al.(2019)Voita, Talbot, Moiseev, Sennrich, and Titov]{voita2019lrp}
E.~Voita, D.~Talbot, F.~Moiseev, R.~Sennrich, and I.~Titov.
\newblock Analyzing multi-head self-attention: Specialized heads do the heavy lifting, the rest can be pruned.
\newblock In \emph{Proceedings of the 57th Annual Meeting of the Association for Computational Linguistics}, pages 5797--5808, 2019.

\bibitem[Xie et~al.(2017)Xie, Girshick, Doll{\'a}r, Tu, and He]{xie2017resnet}
S.~Xie, R.~Girshick, P.~Doll{\'a}r, Z.~Tu, and K.~He.
\newblock Aggregated residual transformations for deep neural networks.
\newblock In \emph{Proceedings of the IEEE conference on computer vision and pattern recognition}, pages 1492--1500, 2017.

\bibitem[Xu et~al.(2015)Xu, Ba, Kiros, Cho, Courville, Salakhudinov, Zemel, and Bengio]{xu2015show-attend-tell}
K.~Xu, J.~Ba, R.~Kiros, K.~Cho, A.~Courville, R.~Salakhudinov, R.~Zemel, and Y.~Bengio.
\newblock Show, attend and tell: Neural image caption generation with visual attention.
\newblock In \emph{International conference on machine learning}, pages 2048--2057. PMLR, 2015.

\bibitem[Yao et~al.(2022)Yao, Wang, and Jin]{yao2022image}
L.~Yao, W.~Wang, and Q.~Jin.
\newblock Image difference captioning with pre-training and contrastive learning.
\newblock In \emph{Proceedings of the AAAI Conference on Artificial Intelligence}, volume~36, pages 3108--3116, 2022.

\bibitem[Yao et~al.(2023)Yao, Zhang, Wang, Hou, Ge, Jiang, and Jin]{yao2023edit}
L.~Yao, Y.~Zhang, Z.~Wang, X.~Hou, T.~Ge, Y.~Jiang, and Q.~Jin.
\newblock Edit as you wish: Video description editing with multi-grained commands, 2023.

\bibitem[Ye et~al.(2023)Ye, Xu, Xu, Ye, Yan, Zhou, Wang, Hu, Shi, Shi, Jiang, Li, Xu, Chen, Tian, Qi, Zhang, and Huang]{ye2023mplugowl}
Q.~Ye, H.~Xu, G.~Xu, J.~Ye, M.~Yan, Y.~Zhou, J.~Wang, A.~Hu, P.~Shi, Y.~Shi, C.~Jiang, C.~Li, Y.~Xu, H.~Chen, J.~Tian, Q.~Qi, J.~Zhang, and F.~Huang.
\newblock mplug-owl: Modularization empowers large language models with multimodality, 2023.

\bibitem[Yu et~al.(2016)Yu, Poirson, Yang, Berg, and Berg]{yu2016modeling_refcoco}
L.~Yu, P.~Poirson, S.~Yang, A.~C. Berg, and T.~L. Berg.
\newblock Modeling context in referring expressions.
\newblock In \emph{Computer Vision--ECCV 2016: 14th European Conference, Amsterdam, The Netherlands, October 11-14, 2016, Proceedings, Part II 14}, pages 69--85. Springer, 2016.

\bibitem[Zhai et~al.(2023)Zhai, Mustafa, Kolesnikov, and Beyer]{zhai2023siglip}
X.~Zhai, B.~Mustafa, A.~Kolesnikov, and L.~Beyer.
\newblock Sigmoid loss for language image pre-training.
\newblock In \emph{Proceedings of the IEEE/CVF International Conference on Computer Vision}, pages 11975--11986, 2023.

\bibitem[Zhao et~al.(2023)Zhao, Cai, Si, Ma, An, Chen, Liu, Wang, Han, and Chang]{zhao2023mmicl}
H.~Zhao, Z.~Cai, S.~Si, X.~Ma, K.~An, L.~Chen, Z.~Liu, S.~Wang, W.~Han, and B.~Chang.
\newblock Mmicl: Empowering vision-language model with multi-modal in-context learning.
\newblock \emph{ArXiv preprint}, abs/2309.07915, 2023.

\bibitem[Zhu et~al.(2023)Zhu, Chen, Shen, Li, and Elhoseiny]{zhu2023minigpt4}
D.~Zhu, J.~Chen, X.~Shen, X.~Li, and M.~Elhoseiny.
\newblock Minigpt-4: Enhancing vision-language understanding with advanced large language models.
\newblock \emph{ArXiv preprint}, abs/2304.10592, 2023.

\bibitem[Zhu et~al.(2020)Zhu, Su, Lu, Li, Wang, and Dai]{zhu2020deformable}
X.~Zhu, W.~Su, L.~Lu, B.~Li, X.~Wang, and J.~Dai.
\newblock Deformable detr: Deformable transformers for end-to-end object detection.
\newblock In \emph{International Conference on Learning Representations}, 2020.

\end{thebibliography}

\newpage

\appendix
\section{Details of R-GAE Explainability Tool}

\subsection{Background: GAE Explainability for Transformer Layers}
\label{subsec:preliminary}
The Generic Attention Explainability (GAE)~\cite{chefer2021gae} is a powerful method to interpret predictions for bi-modality Transformer-based architectures. It has the advantage of acquiring the relevance map from two arbitrary layers in the Transformer through propagation.
Essentially, the GAE method generates a relevance map $\mathbf{\bar{A}}$ for each self-attention layer or cross-attention layer by integrating raw attention maps and gradients. Then it aggregating the relevance maps of all layers into a overall single map $\mathbf{R}$. Formally, denote a Transformer architecture as $\phi$, its attention map of each layer as $\mathbf{A}$, the input modality tokens as $I \in \mathbb{R}^{N \times d}$ and the output predict class as $y$. We aim to visualize the relevance map  $\mathbf{R}_{y\rightarrow{}I}\in \mathbb{R}^{N}$ from class $y$ to input tokens $I$. Take the self-attention layer as an example, the relevance map $\mathbf{\bar{A}}$ for each layer and the propagation of  final map $\mathbf{R}_{y\rightarrow{}I}$ are termed as:
\noindent
 \begin{align}
   \label{eq:gae}
   & 
    \mathbf{\bar{A}} = \mathbb{E}_h ((\nabla \mathbf{A} \odot \mathbf{A})^+),
\end{align}
\noindent
 \begin{align}
   \label{eq:gae} &
    \mathbf{R} = \mathbf{R}+\mathbf{\bar{A}} \cdot \mathbf{R},
\end{align}

where each layer's attention map $\mathbf{A}$ can be obtained through a forward pass and the related gradient $ \nabla \mathbf{A} := \frac{\partial \mathcal{\phi}(y)}{\partial \mathbf{A}}$ can be cached during a backward pass. 
$\odot$ is the Hadamard product, $(\cdot)^+$ represents the operation of setting negative values to 0, and $\mathbb{E}_h$ is the mean across the attention heads dimension. 
The overall map $\mathbf{R}$ is initialized as the identity matrix with the intuition that each input token's relevance score is identical in the beginning. The propagation Formula~\ref{eq:gae} updates the  $\mathbf{R}$ from a start layer $L_{s}$ to an  end layer $L_{e}$ ($L_{e}>L_{s}$ ) in the Transformer. The cross-attention propagation is similar, which maintains two relevance matrices for two modalities and updates them through the layer interaction. Please refer to the details of the propagation formula across cross-attention layer from the original paper.

\subsection{R-GAE Propagation for MLLMs}

The traditional GAE map is designed for a classification task with the special $CLS$ token. We adapt it to MLLM architectures and propose the R-GAE explainability tool.
As Figure~\ref{fig:supp_overall} shows, a typical MLLM architecture comprises a Vision Transformer (ViT) $\phi_v$ to acquire patch-level visual representations $\mathcal{I} \in \mathbb{R}^{N \times d_{I}}$ 
 (containing $N$ patches), a projector $\phi_p$ to transform visual representations into the textual embedding space as $\mathcal{Q}$, and an LLM $\phi_{t}$ that handles both vision and instruction tokens to output hidden states $\mathcal{T}  \in \mathbb{R}^{L \times d_{T}}$ and generate responses $Y = \{y_1, y_2, \dots, y_L\}$.  We summarize widely adopted projectors into two branches:

\textit{Non-compressive Projectors} maintain the number of patch tokens $N$ and only transform the visual embedding dimension to match the dimension of the LLM, as exemplified by the linear projector~\cite{liu2023llava}. The projected visual tokens can be denoted as $\mathcal{Q}  \in \mathbb{R}^{N \times d_{T}}$. 

\textit{Compressive Projectors} reduce the number of patch tokens  $N$ to a specified lesser number $M$ ($M<N$), conserving training resources. For instance, QFormer~\cite{li2023blip2} learns pre-defined query tokens to compress original visual tokens. These compressed query tokens  $\mathcal{Q}  \in \mathbb{R}^{M \times d_{T}}$ 
 are then fed into the LLM providing vision information.

 We initialize three GAE relevance maps including a Text-to-Patch map as $\mathbf{R}_{\mathcal{T}\rightarrow{}\mathcal{I} } $, a Text-to-Query map as $\mathbf{R}_{\mathcal{T}\rightarrow{}\mathcal{Q} } $, and a Query-to-Patch map as $\mathbf{R}_{\mathcal{Q}\rightarrow{}\mathcal{I} } $. 
As Figure~\ref{supp_overall} depicts, given an image and an instruction (e.g., ``\textit{Please describe the image with a concise sentense}''), an MLLM will generate a textual description $Y = \{y_1, y_2, \dots, y_L\}$ referring to the visual information. During the generation step $t$, we can cache the attention map $\mathbf{A}_v$, $\mathbf{A}_p$, $\mathbf{A}_t$ across the ViT, the projector and the LLM during a forward pass. Then specifying a word class $\widehat{y}_t$ as the target predict,  we can get the gradients  $ \nabla \mathbf{A}_t$, $ \nabla \mathbf{A}_p$, $ \nabla \mathbf{A}_v$  in each module through a backward pass. The LLM module in MLLMs substantially contains self-attention layers, therefore, we can propagate the $\mathbf{R}^t_{\mathcal{T}\rightarrow{}\mathcal{Q} } \in \mathbb{R}^{1\times M}$ according to Formula~\ref{eq:gae} from LLM's first layer to its last layer. The QFormer-type projector consisting of self-attention and cross-attention layers can also be propagated similarly to get $\mathbf{R}^t_{\mathcal{Q}\rightarrow{}\mathcal{I} } \in \mathbb{R}^{M\times N}$. Subsequently, the overall text-to-patch relevance map can be obtained by matrix multiplication of text-to-query and query-to-patch maps:
\begin{align}
    \mathbf{R}^t_{\mathcal{T}\rightarrow{}\mathcal{I} }&=
    \mathbf{R}^t_{\mathcal{T}\rightarrow{}\mathcal{Q} } 
    \times \mathbf{R}^t_{\mathcal{Q}\rightarrow{}\mathcal{I} }
\end{align}

For a complete sentence $Y$, we integrate the GAE relevance maps from each time step $t$ by averaging them to obtain the overall visual relevance related to a factual sentence. The final three maps are formulated as followings, in which 
$\mathbf{R}_{\mathcal{T}\rightarrow{}\mathcal{I} }  \in \mathbb{R}^{1\times N}$,
$\mathbf{R}_{\mathcal{T}\rightarrow{}\mathcal{Q} }  \in \mathbb{R}^{1 \times M}$, and 
$\mathbf{R}_{\mathcal{Q}\rightarrow{}\mathcal{I} }  \in \mathbb{R}^{M\times N}$.
\noindent
\begin{align}
    \label{eq:final_gae_maps}
    \mathbf{R}_{\mathcal{T}\rightarrow{}\mathcal{Q} } &= \frac{1}{L} \sum_{t=1}^L \mathbf{R}^t_{\mathcal{T}\rightarrow{}\mathcal{Q} },
    &\mathbf{R}_{\mathcal{Q}\rightarrow{}\mathcal{I} } &= \frac{1}{L} \sum_{t=1}^L \mathbf{R}^t_{\mathcal{Q}\rightarrow{}\mathcal{T} },
    &\mathbf{R}_{\mathcal{T}\rightarrow{}\mathcal{I} } &= \frac{1}{L} \sum_{t=1}^L \mathbf{R}^t_{\mathcal{T}\rightarrow{}\mathcal{I} }
\end{align}

For non-compression projectors maintaining the number of original patches, such as linear layers, the Query-to-Patch map is an identity mapping based on the one-to-one correspondence between queries and patches. Consequently, the Query-to-Image map visualizes the original image consisting of 576 patches. The Text-to-Query map is obtained in the same manner as in the QFormer, which propagates from the R-GAE maps in the Language Model (LLM).

For the \avgpool projector in the \model method, a 2D spatial down-sampling mapping is constructed from the original tokens to the compressed tokens. For an operation window with kernel size $K$, the merged token is assigned a relevance score equal to $1/K^2$ of the sum of the relevance scores of each raw token within the window. The corresponding Query-to-Patch map can be calculated using this simple mapping rule. Similar to the QFormer, the Text-to-Query map is obtained from the LLM layers.

\section{Comparison between R-GAE and Raw Attention Maps}
The R-GAE map offers two advantages over raw attention maps: (i) it demonstrates better explainability~\cite{chefer2021gae} by integrating both attention maps and gradients, and (ii) it can track token relevance from a target layer (e.g., output textual tokens) to the first layer (e.g., original patch tokens). In contrast, the attention map commonly used from the last layer of the Large Language Model (LLM) can only show the relevance of mixed tokens in that layer, where the tokens related to the vision input position or the output word position have incorporated the semantics of other tokens through attention operation in previous layers.

\begin{figure}[tbp]
    \centering
    \includegraphics[width=\linewidth]{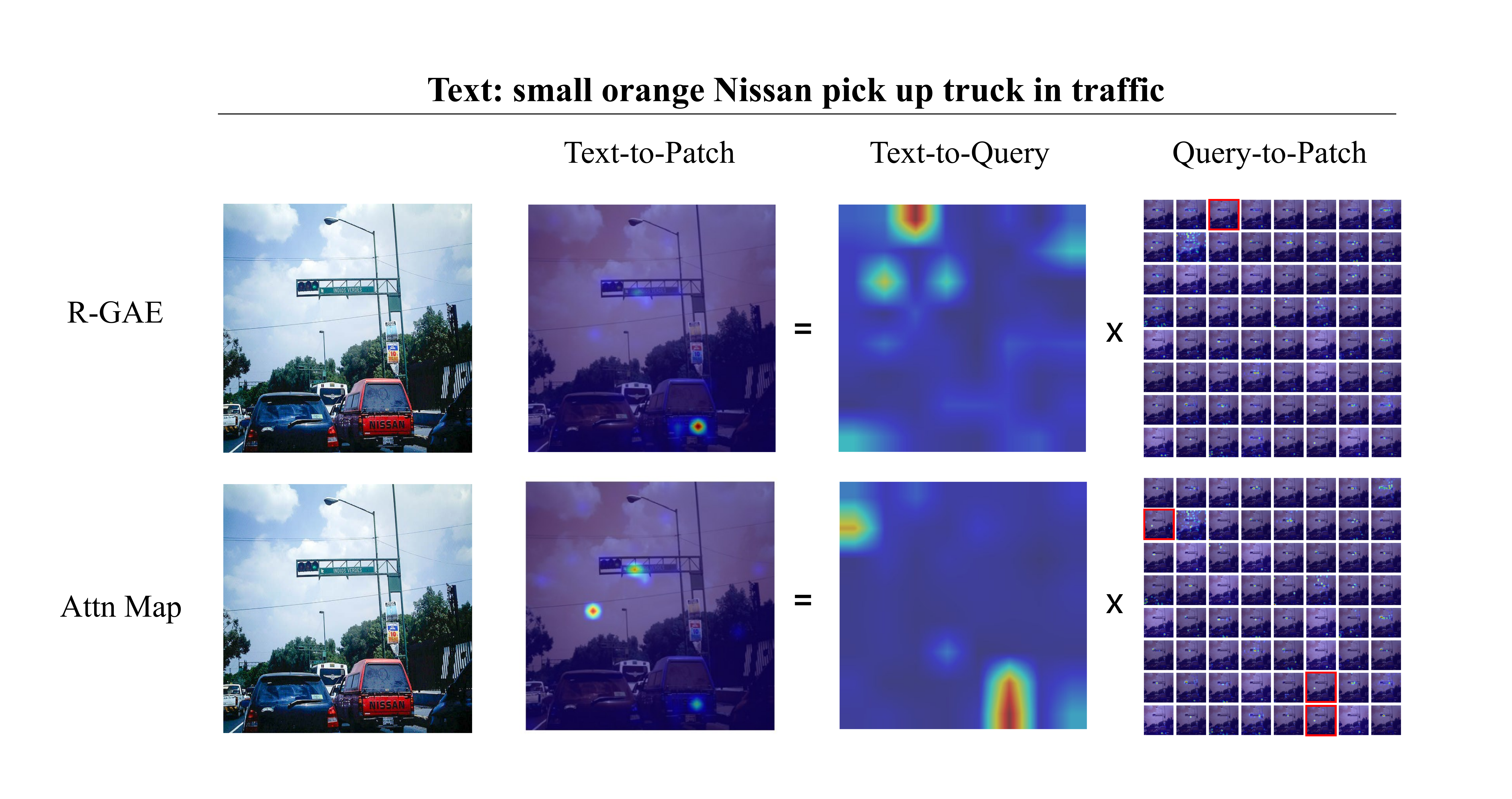}
    \caption{Comparison between the R-GAE and raw attention map explainability on the same case from the QFormer projector, which compresses 576 vision tokens to 64 query tokens.
    }
    \label{fig:supp_attn}
\end{figure}

Figure~\ref{fig:supp_attn} visualizes the R-GAE relevance maps and the raw attention maps for a comparative analysis. The Query-to-Patch map of raw attention is obtained from the last cross-attention layer in the QFormer, while the Text-to-Query map of raw attention is derived from the last layer in the LLM. By visualizing the same model and image-text pair, it becomes evident that R-GAE provides a more interpretable representation of the inner vision-language alignment of an MLLM.
In contrast, the raw attention map highlights an unrelated visual patch, such as the sky, which introduces an additional error in the explainability procedure when analyzing semantic alignment. An in-depth analysis reveals that the error in the raw attention map primarily originates from the Text-to-Query map of the last LLM layer. This can be attributed to the fact that the LLM consists of 32 self-attention layers, and the relevance among query tokens and text tokens in the last layer has deviated due to the fusion of semantics from other tokens in previous layers.
On the other hand, the Query-to-Patch map exhibits relatively similar characteristics to the R-GAE map. This similarity can be explained by the architecture of the QFormer, which only employs a single cross-attention layer, thus minimizing the influence of token fusion across layers for raw attention.


\section{Training Hyper-parameters}
\label{sec:hyper-parameters}
\paragraph{Architecture of Used Projectors.}
\begin{enumerate}
\item C-Abstractor comprises 3-layer ResNet blocks~\cite{xie2017resnet}, the adaptive average pooling and another 3-layer ResNet blocks.
\item D-Abstractor leverages Deformable Attention~\cite{zhu2020deformable} to replace the vanilla attention and conduct well-designed initialization of query tokens. We adopt a two-layer D-Abstractor.
\item QFormer is a two-layer BERT~\cite{devlin2019bert} architecture same as the the BLIP-2~\cite{li2023blip2} and we load the BLIP-2 pre-training weights as an initialization.
\item Linear projector is a two-layer MLP with the GELU activation same as the LLaVA v1.5~\cite{liu2023llava15}.
\item \avgpool is parameter-free, we utilize a two-layer MLP as the linear projector to map the vision feature dimension to the LLM's.
\end{enumerate}

\paragraph{Training Parameters.}
Our experiments are conducted under two primary training settings. The main experiments are built on the LLaVA v1.5 framework, as shown in Table~\ref{tab:supp_llava-para}. The generalization experiments are constructed using a more lightweight setup that involves only the instruction tuning stage, referring to the PRISM~\cite{prismatic} approach. Specific training hyperparameters are detailed in Table~\ref{tab:supp_prism-para}.

\begin{table}[h]
\centering
\caption{Hyper-parameters of main experiments.}
\begin{tabular}{l|cc}
\toprule
\textbf{Hyperparameter} & \textbf{Pretrain} & \textbf{Finetune} \\ \midrule
batch size & 256 & 128 \\
lr & 1e-3 & 2e-5 \\ 
lr schedule & \multicolumn{2}{c}{cosine decay} \\ 
lr warmup ratio & \multicolumn{2}{c}{0.03} \\ 
weight decay & \multicolumn{2}{c}{0} \\ 
epoch & \multicolumn{2}{c}{1} \\ 
optimizer & \multicolumn{2}{c}{AdamW} \\ 
DeepSpeed stage & 2 & 3 \\ \bottomrule
\end{tabular}

\label{tab:supp_llava-para}
\end{table}

\begin{table}[h]
\centering
\caption{Hyper-parameters of generalization experiments.}
\begin{tabular}{l|c}
\toprule
\textbf{Hyperparameter} & \textbf{Value} \\ \midrule
Batch Size & 128 \\ 
Max Gradient Norm & 1.0 \\ 
Weight Decay & 0.1 \\ 
Learning Rate & 2e-5 \\ 
Optimizer & AdamW \\ 
Scheduler & Warmup \& Cosine Decay \\ 
Warmup Ratio & 0.03 \\ \bottomrule
\end{tabular}

\label{tab:supp_prism-para}
\end{table}

\section{More R-GAE Relevance Maps}
\begin{figure}[tbp]
    \centering
    \begin{subfigure}[b]{\textwidth}
        \centering
        \includegraphics[width=\linewidth]{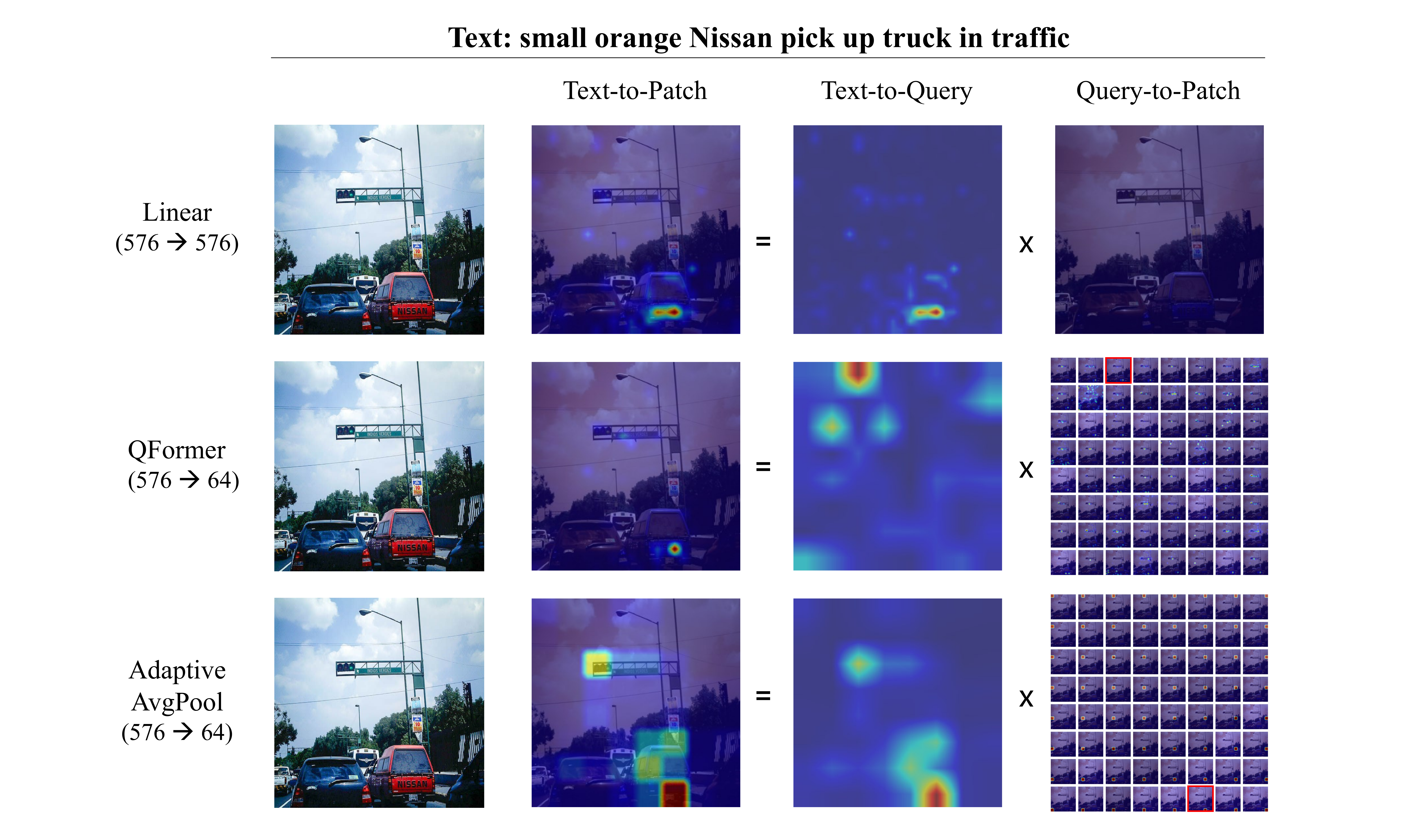}
    \caption{R-GAE maps related to the generated text ``small orange Nissan pick up truck in traffic''. In this case, all projectors reserve the effective visual representation and translate it to the LLM. Specifically, the Qformer-based MLLM attends to the query indexing $(0,2)$ which highlights the ``Nissan'' semantics on the image. This indicates extracting effective visual semantic concepts in the first abstraction by the QFormer is important for the traditional compressive projectors.}
    \end{subfigure} \\
    \begin{subfigure}[b]{\textwidth}
        \centering
        \includegraphics[width=\linewidth]{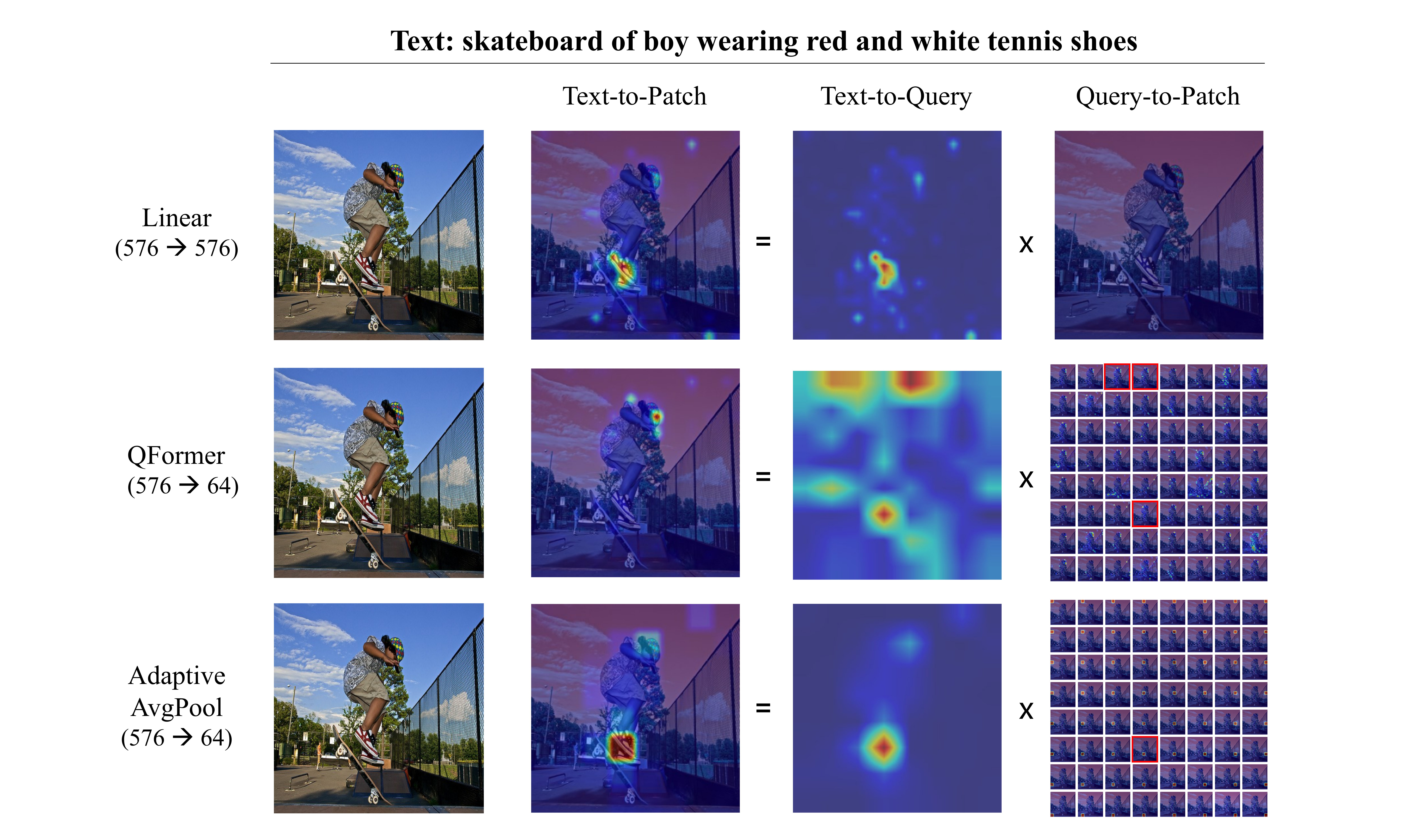}
    \caption{R-GAE maps related to the generated text ``skateboard of boy wearing red and white tennis shoes'' are shown in Figure~\ref{fig:supp_model_r-gae}. In this case, the QFormer-based MLLM fails to attend to the relevant patches with the ``red and white tennis shoes'' attributes. In contrast, both the linear projector and the \avgpool highlight the correct patches.}
    \end{subfigure}
     \caption{ Visualization of additional R-GAE relevance maps. The linear projector is non-compressive, while the QFormer and Adaptive Average Pooling (\avgpool) compress the original 576 vision tokens to 64. For better visualization, the highlighted query tokens from the text are framed in red.}
    \label{fig:supp_model_r-gae}
\end{figure}

Figure~\ref{fig:supp_model_r-gae} presents additional visualized cases of the R-GAE relevance map across different projectors.

\section{Broader Impacts}
\label{sec:broader-impacts}
Our work utilizes off-the-shelf frozen LLMs, which means it shares some of their intrinsic drawbacks, such as generating hallucinated, ungrounded text or biased outputs. We mitigate these issues by enhancing the model’s grounding in both visual and instruction inputs. Additionally, our training dataset includes 40K examples of safety data sourced from ShareGPT, instructing the models to refuse responses to toxic, inappropriate, or otherwise unsafe inputs. However, we do not recommend applying our models to any downstream applications without a prior assessment of safety and fairness specific to that application.

\end{document}